\documentclass[runningheads]{llncs}

\usepackage{eccv}

\usepackage{eccvabbrv}

\usepackage{graphicx}
\usepackage{booktabs}
\usepackage{amsmath}
\usepackage{amssymb}

\usepackage{color}
\usepackage{amssymb}
\usepackage{lipsum}
\usepackage{algorithm}
\usepackage{algpseudocode}
\usepackage{xcolor}
\usepackage{tabularx}
\usepackage{multirow}
\usepackage{enumitem}
\usepackage{bbm}
\usepackage{wrapfig}
\usepackage{setspace}
\usepackage{footnote}

\usepackage{subcaption}
\usepackage{colortbl} 
\usepackage{pifont}  
\usepackage{cite}  
\usepackage{mathtools}  
\usepackage[font=small]{caption}  
\usepackage{blindtext}  
\usepackage{stmaryrd} 
\usepackage{mathtools}
\usepackage{subfiles}
\usepackage{afterpage}

\usepackage{algorithm}
\usepackage{algpseudocode}
\usepackage{adjustbox}

\usepackage{afterpage}
\usepackage{xcolor,pifont}
\usepackage{makecell}

\usepackage[accsupp]{axessibility}  

\usepackage[pagebackref,breaklinks,colorlinks,citecolor=eccvblue]{hyperref}

\usepackage{orcidlink}

\definecolor{amethyst}{rgb}{0.6, 0.4, 0.8}
\definecolor{grey}{rgb}{0.93, 0.93, 0.93}
\newcommand{\ccol}{\cellcolor{grey}}
\newcommand*\colourcheck[1]{%
  \expandafter\newcommand\csname #1check\endcsname{\textcolor{#1}{\ding{51}}}%
}

\colourcheck{blue}
\colourcheck{green}
\colourcheck{red}
\newcommand{\cmark}{\ding{51}}%
\newcommand{\xmark}{\ding{55}}%
\newcommand{\expnum}[2]{{#1}\mathrm{e}{-#2}}

\def\ie{\emph{i.e.}}
\def\eg{\emph{e.g.}}
\def\etal{\emph{et al.}}

\definecolor{bright_red}{rgb}{0.97, 0.9, 0.9}
\definecolor{blk}{rgb}{0, 0, 0}
\definecolor{grn}{rgb}{0, 0.6, 0}
\definecolor{mgt}{rgb}{0.8, 0.1, 0.8}
\definecolor{darkblue}{rgb}{0.2, 0.2, 0.8}
\definecolor{lblue}{rgb}{0.2, 0.2, 1.0}
\definecolor{orange}{rgb}{0.8, 0.2, 0}
\definecolor{goldenrod}{rgb}{0.85, 0.65, 0.13}

\makeatletter
\newcommand{\multiline}[1]{%
  \begin{tabularx}{\dimexpr\linewidth-\ALG@thistlm}[t]{@{}X@{}}
    #1
  \end{tabularx}
}
\makeatother

\usepackage[capitalize]{cleveref}
\crefname{section}{Sec.}{Secs.}
\Crefname{section}{Section}{Sections}
\Crefname{table}{Table}{Tables}
\crefname{table}{Tab.}{Tabs.}

\begin{document}

\title{Efficient and Versatile Robust Fine-Tuning of Zero-shot Models}

\titlerunning{Efficient and Versatile Robust Fine-Tuning of Zero-shot Models}

\author{Sungyeon Kim\inst{1}\orcidlink{0000-0002-6919-4822} \quad
Boseung Jeong\inst{1}\orcidlink{0000-0001-9382-3396} \quad
Donghyun Kim\inst{2}\orcidlink{0000-0002-7132-4454}  \quad
Suha Kwak\inst{1,3}\orcidlink{0000-0002-4567-9091}}

\authorrunning{S.~Kim \etal}

\institute{
Department of Computer Science and Engineering, POSTECH, Korea
\and Department of Artificial Intelligence, Korea University, Korea
\and Graduate School of Artificial Intelligence, POSTECH, Korea \\
\inst{1}\email{\{sungyeon.kim,boseung01,suha.kwak\}@postech.ac.kr, 
\inst{2}d\_kim@korea.ac.kr}\\
{\tt\small
\url{http://cvlab.postech.ac.kr/research/R-Adapter}}
}

\maketitle

\begin{abstract}
Large-scale image-text pre-trained models enable zero-shot classification and provide consistent accuracy across various data distributions. 
Nonetheless, optimizing these models in downstream tasks typically requires fine-tuning, which reduces generalization to out-of-distribution (OOD) data and demands extensive computational resources. 
We introduce Robust Adapter~(R-Adapter), a novel method for fine-tuning zero-shot models to downstream tasks while simultaneously addressing both these issues. Our method integrates lightweight modules into the pre-trained model and employs novel self-ensemble techniques to boost OOD robustness and reduce storage expenses substantially.
Furthermore, we propose MPM-NCE loss designed for fine-tuning on vision-language downstream tasks. It ensures precise alignment of multiple image-text pairs and discriminative feature learning. By extending the benchmark for robust fine-tuning beyond classification to include diverse tasks such as cross-modal retrieval and open vocabulary segmentation, we demonstrate the broad applicability of R-Adapter. Our extensive experiments demonstrate that R-Adapter achieves state-of-the-art performance across a diverse set of tasks, tuning only 13\% of the parameters of the CLIP encoders.
\keywords{robust fine-tuning; parameter-efficient fine-tuning; self-ensemble}
\end{abstract}
\section{Introduction}
\label{sec:intro}

The emergence of large-scale models pre-trained jointly on image and text data~\cite{clip, align, li2022blip} brings a paradigm shift in the field of computer vision. By aligning the embeddings of extensive image-text pairs, these models enable zero-shot inference and show a remarkable ability to generalize across diverse data distributions. Despite their impressive performance in a zero-shot context, they do not measure up to supervised learning models~\cite{radford2021learning, wiseft}, necessitating fine-tuning to unlock their full capabilities. While conventional full fine-tuning enhances task-specific performance, it introduces two major challenges:
\textbf{1)} Full fine-tuning compromises the ability of the model to generalize to out-of-distribution (OOD) data, crucial for real-world applications where data variability is unpredictable. \textbf{2)} It demands substantial computational resources, memory, and storage, which is impractical given the growing size of large pre-trained models.

\begin{figure}[!t]
\centering
\includegraphics[width = 0.99\linewidth]{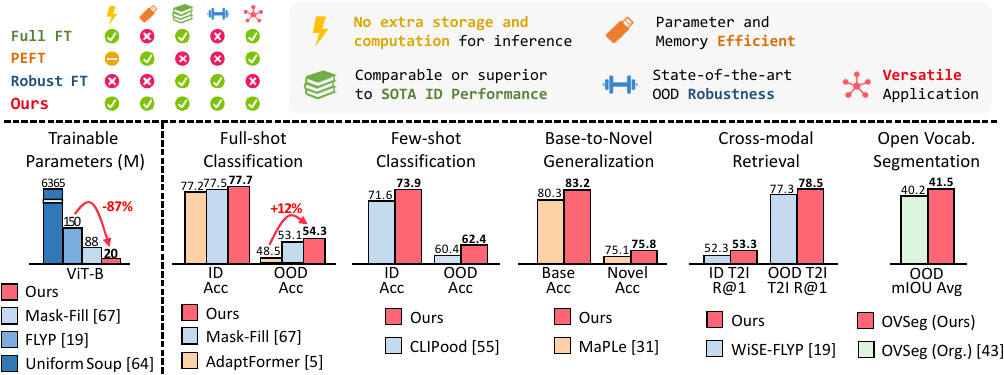}
\caption{We present Robust Adapter (R-Adapter), which combines the strengths of robust fine-tuning and parameter-efficient fine-tuning~(PEFT). R-Adapter improves parameter and memory efficiency compared to existing robust fine-tuning (\eg, Mask-fill~\cite{maskfill}, ModelSoup~\cite{modelsoup}) while being more robust compared to existing PEFT (\eg, AdaptFormer~\cite{adaptformer}, MaPLe~\cite{maple}). Unlike most of existing robust fine-tuning, our method can apply to a wide range of tasks, and consistently outperforms current best methods on diverse tasks in both in-distribution (ID) and out-of-distribution (OOD).} 
\label{fig:thumbnail}
\end{figure}

Recently, several fine-tuning approaches have been proposed to address these challenges. \emph{Robust fine-tuning}~\cite{flyp,lpft,maskfill, wiseft,modelsoup} aims to fine-tune zero-shot models while preserving their robustness to OOD, and \emph{Parameter-Efficient Fine-Tuning} (PEFT)~\cite{coop, houlsby2019parameter,pfeiffer2020adapterfusion,adaptformer,repadapter,hu2021lora, maple} updates only a small set of parameters while keeping pre-trained parameters frozen. However, each approach addresses only one of the challenges while still falling short on the other. As shown in Fig.~\ref{fig:thumbnail}, existing robust fine-tuning methods still require tuning the entire model, making training expensive. Moreover, they have only targeted classification tasks, thus often training solely image encoder and excluding zero-shot inference capabilities from the model. On the other hand, PEFT significantly lags in performance compared to robust fine-tuning under distribution shifts.
Their critical shortcomings highlight the need for new fine-tuning methods that \emph{simultaneously address both challenges tackled by robust fine-tuning and PEFT separately}.

This paper presents \textbf{Robust Adapter (R-Adapter)}, a novel fine-tuning method for improving the robustness of PEFT while enhancing the efficiency of robust fine-tuning. 
Building upon the adapter-tuning approach~\cite{adaptformer, repadapter}, where extra lightweight modules are added to a pre-trained model, R-Adapter incorporates novel self-ensemble strategies to enhance OOD robustness.

We take inspiration from the robustness gain observed when averaging multiple models in the weight-space~\cite{modelsoup, wiseft}, yet implement this strategy within a single model via a unique way. 
This approach strikes a good balance between task-specific performance and robustness against distribution shifts, and at the same time significantly reduces storage costs. 
Specifically, R-Adapter achieves this through three self-ensemble techniques. It randomly drops the adapter module, thereby dynamically generating and ensemble different subnetworks combining both the adapter and pre-trained layers in various configurations. 
Additionally, we accumulate adapter weights to form a temporal ensemble that captures all models derived throughout the learning process.
Moreover, by re-scaling the weights of the adapter and integrating it into the pre-trained layer via re-parametrization, we enable a seamless linear interpolation between the weights of the pre-trained and fine-tuned models without two separate models.


Additionally, we propose the \textbf{Multi-Positive Margin NCE (MPM-NCE)} loss function designed for effective fine-tuning on vision-language downstream tasks. These tasks often involve intricate relations where multiple images can correspond to the same text, and vice versa. Unlike traditional contrastive loss, \ie, InfoNCE~\cite{Sohn_nips2016,oord2018representation}, which takes single positive pairs and therefore often leads to semantic mismatches in these relations, MPM-NCE accounts for multiple positive pairs and thus promotes more precise alignment across various image-text pairs.
Moreover, MPM-NCE introduces an angular margin to penalize negative pairs, enabling the model to learn highly discriminative features critical for downstream tasks. Consequently, the proposed loss leads to significant improvement in task-specific performance, offering benefits in both ID and OOD contexts.

Our method enables zero-shot inference after fine-tuning, extending its applicability beyond image classification tasks to a wide range of applications. To show its versatility, we present a new evaluation benchmark for robust fine-tuning that includes five tasks: image classification tasks under three scenarios, cross-modal retrieval, and open-vocabulary segmentation.
Extensive experiments demonstrate that our method achieves superior performance under distribution shift while using fewer parameters compared to existing robust fine-tuning and PEFT methods. The main contribution of this paper is four-fold:
\begin{itemize}[leftmargin=*, topsep=1mm]
\item We introduce an efficient and versatile framework for robust fine-tuning that incorporates the strengths of both PEFT and robust fine-tuning. To the best of our knowledge, it is the first method to make the best of both worlds.
\item We propose \textbf{R-Adapter} with self-ensemble techniques enabling weight-space ensemble using a single model with adapters. These techniques enhance robustness while reducing storage costs, as it does not need multiple models.
\item We develop \textbf{MPM-NCE} loss tailored for fine-tuning,
utilizing multiple positive pairs and introducing an angular margin. This loss ensures precise alignment of multiple image-text pairs and discriminative feature learning. 
\item For the first time, we extend the benchmark for robust fine-tuning beyond image classification to include tasks such as cross-modal retrieval and open vocabulary segmentation, allowing us to assess the broad applicability. As shown in Fig.~\ref{fig:thumbnail}, our method achieves state-of-the-art performance on diverse tasks while tuning only 13\% of CLIP encoder parameters.
\end{itemize}

\section{Related Work}
\label{sec:relate}

\noindent\textbf{Robust Fine-tuning.} 
In the conventional practice of leveraging pre-trained models, linear probing or full fine-tuning are commonly used methods for fine-tuning pre-trained models. 
Kumar et al.~\cite{lpft} show that while fine-tuning achieves higher accuracy on in-distribution (ID) data, it can distort pre-trained knowledge, reducing out-of-distribution (OOD) accuracy.
To mitigate this, a two-step process involving linear probing followed by full fine-tuning has been suggested. 
Following this paradigm, ensembling-based robust fine-tuning approaches have been proposed in~\cite{modelsoup, wiseft}. WiSE-FT~\cite{wiseft} ensembles weights of pre-trained fine-tuned models, improving accuracy on both ID and OOD data. FLYP~\cite{flyp} reuses the same contrastive formulation from pre-training for fine-tuning. Mask-Fill~\cite{maskfill} promotes consistency between fine-tuned and pre-trained models on counterfactual samples. However, these require full fine-tuning or additional forward/backward passes, leading to high memory and computational demands. Given the substantial size of the foundation models, we aim to develop efficient and fast adaptation methods while improving ID and OOD accuracy.
While earlier work primarily focuses on image classification tasks, we extend our investigation to a broader range of tasks, showing the versatility of our approach.

\noindent\textbf{Parameter-Efficient Fine-Tuning.}
In the context of ever-growing model sizes, fine-tuning large-scale models for various downstream tasks presents a significant challenge, demanding substantial memory and computational resources. To solve this issue, PEFT has been proposed~\cite{jia2022visual,hu2021lora,multi_domain,houlsby2019parameter,pfeiffer2020adapterfusion,he2021towards,adaptformer,scalingshifting}. These methods selectively update a limited portion of trainable parameters,
while keeping pre-trained parameters frozen.
The concept of low-rank adaptation~\cite{hu2021lora,qlora} is introduced to provide an approximation for the parameter update. Several methods only update additional learnable tokens during fine-tuning~\cite{lester2021power,li2021prefix,wang2022learning,smith2022coda,coop,cocoop} while freezing all the parameters. It is feasible to incorporate lightweight adapter modules~\cite{houlsby2019parameter,pfeiffer2020adapterfusion,adaptformer,kim2023universal,repadapter} and only update these modules during fine-tuning. 
However, na\"ively using additional learnable tokens and adapters could increase inference costs.  RepAdapter~\cite{repadapter} proposes a re-parameterization trick for adapters and achieves zero additional cost during inference. 
We propose R-Adapter which employs PEFT for efficient adaptation of large models to diverse downstream tasks and enhancing ID performance and OOD robustness.

\noindent\textbf{{Contrastive Learning.}} 
Contrastive loss has been explored in various fields including self-supervised learning~\cite{nonparametricid,moco,chen2020improved}, vision-language pre-training~\cite{clip,align}, supervised learning~\cite{khosla2020supervised,graf2021dissecting}, metric learning~\cite{Chopra2005,Sohn_nips2016, kim2021embedding}, image captioning~\cite{dai2017contrastive,sarto2023positive}, etc. Contrastive learning trains a model to differentiate between similar (positive) and dissimilar (negative) data sample pairs.  
Recently, contrastive learning on web-crawled image-caption data~\cite{clip} has shown significant gains in zero-shot classification and domain robustness. 
FLYP~\cite{flyp} proposes a fine-tuned approach using the same contrastive learning formulation for image classification with class prompt templates, but this can cause class collision issues between the same positive classes. 
To address this, we introduce MPM-NCE which leverages multiple positive relations, considering the characteristics of downstream tasks.


\section{Proposed Method}
\label{sec:method}
Our method is compatible with various zero-shot models~\cite{align,li2022blip}, but our research primarily centers on the most renowned model, CLIP~\cite{clip}.
In this section, we first revisit the CLIP encoders~\cite{clip} and their pre-training scheme (Sec.~\ref{sec:CLIP}). Next, we define the problem setup (Sec.~\ref{sec:problem}). Then our R-Adapter (Sec.~\ref{sec:R-Adapter}) and MPM-NCE loss (Sec.~\ref{sec:MPM_NCE}) are introduced.
\subsection{Preliminary}
\label{sec:CLIP}

\noindent\textbf{CLIP Encoders.} CLIP consists of two encoders for extracting features from image and text, respectively. Each encoder is composed of a series of Transformer layers~\cite{transformer}, each of which consists of Multi-Head Attention (MHA), Layer Normalization (LN), and Feed-Forward Network (FFN). Specifically, the $l$-th Transformer layer is formulated as follows:
\begin{equation}
\begin{aligned}
    \Bar{X_l} &= \textrm{MHA}(\textrm{LN}(X_{l-1})) + X_{l-1}, \\
    X_l &= \textrm{FFN}(\textrm{LN}(\Bar{X_l})) + \Bar{X_l}.  
    \label{eq:Transformer_layer}
\end{aligned}
\end{equation}
MHA involves $k$-head self-attention operations on queries, keys, and values, achieved via independent linear projections of the input; it is formulated by
\begin{equation}
\begin{aligned}
    \textrm{MHA}(X) &= [\textrm{Attn}^1(X), ..., \textrm{Attn}^k(X)]W_O,\\
    \textrm{Attn}^i(X) &= \textrm{softmax}\big((XW_{Q}^{i})(XW_{K}^{i})^{\top}/{\sqrt{d_h}} \big)(XW_{V}^{i}), 
    \label{eq:MHA}
\end{aligned}
\end{equation}
where $[\cdot,\cdot]$ denotes concatenation, and $d_h$ is set to $d/k$. 
$W_{Q}^{i}\in\mathbb{R}^{d\times d_h}$, $W_{K}^{i}\in\mathbb{R}^{d\times d_h}$, $W_{V}^{i}\in\mathbb{R}^{d\times d_h}$ and $W_{O}\in\mathbb{R}^{d\times d}$ are linear projection matrices.
FFN consists of two linear layers with a non-linear layer in between:
\begin{equation}
    \textrm{FFN}(X) = \sigma(XW_1+b_1)W_2 + b_2,
    \label{eq:FFN}
\end{equation}
where $W_1\in\mathbb{R}^{d\times4d}$, $W_2\in\mathbb{R}^{4d\times d}$, $b_1 \in \mathbb{R}^{4d}$, and $b_2 \in \mathbb{R}^d$ are the respective linear projection weights and biases; $\sigma(\cdot)$ denotes the GELU function.

\vspace{0.5mm}
\noindent\textbf{Contrastive Learning.}
The CLIP encoders are trained to predict which text descriptions correspond to a given set of images and vice versa. 
This is achieved through contrastive learning using the InfoNCE loss~\cite{oord2018representation}, which forces image embeddings and their corresponding text embeddings to be close to each other and farther away from other text embeddings in a batch.
Let $f(\cdot)$ and $g(\cdot)$ be the CLIP encoders for image and text, respectively.
Given a batch with $B$ image-text pairs $\mathcal{B} =\big\{(I_1,T_1), ..., (I_B,T_B)\big\}$, the loss function is formulated by
\begin{equation}
\begin{aligned}
    \mathcal{L}(\mathcal{B}) = &-\sum_{i=1}^{B}\Bigg(\log\frac{e^{f_i \cdot g_i/\tau }}{\sum_{j=1}^{B}e^{f_i \cdot g_j/\tau }} 
    +\log\frac{e^{f_i\cdot g_i/\tau }}{\sum_{j=1}^{B}e^{f_j\cdot g_i/\tau}}\Bigg),
    \label{eq:InfoNCE_Loss}    
\end{aligned} 
\end{equation}
where $f_i = \frac{f(I_i)}{||f(I_i)||_2}$, $g_i = \frac{g(T_i)}{||g(T_i)||_2}$, $\tau$ denotes a learnable temperature parameter.


\subsection{Problem Setup}
\label{sec:problem}
Our objective is to efficiently fine-tune a vision-language pre-trained model for various downstream tasks while preserving its inherent out-of-distribution (OOD) generalization capability.
While most existing robust-fine tuning methods are limited to classification tasks~\cite{wiseft, maskfill}, we \emph{broaden the scope to robustly fine-tune the models for diverse downstream tasks} such as image classification, cross-modal retrieval, and open-vocabulary segmentation.

Given an image-text pre-trained model, the goal is its adaptation using an in-distribution~(ID) training dataset $\mathcal{D}_{\mathcal{I}}=\{(I_i, T_i)\}_{i=1}^{n}$ for the target downstream task, where $I$ denotes an image and $T$ is a text description corresponding to the image. 
Concurrently, we aim to enhance the performance of the model on an OOD test dataset $\mathcal{D}_{\mathcal{O}}=\{(I_j, T_j)\}_{j=1}^{m}$.
The ID and OOD datasets, $\mathcal{D}_{\mathcal{I}}$ and $\mathcal{D}_{\mathcal{O}}$, are sampled from different probability distributions, $p_{\mathcal{I}}(I,T)$ and $p_{\mathcal{O}}(I,T)$, respectively, exhibiting distribution shift when $p_{\mathcal{I}}(I,T)\neq p_{\mathcal{O}}(I,T)$.
In classification tasks, $T$ represents a text description of the target class which is constructed by sampling from a set of predefined templates (\eg, ``\texttt{a photo of a \{class\}}'')~\cite{flyp, clip}.
For other vision-language tasks, $T$ could be one of the captions associated with the image $I$~\cite{Mscoco, Flickr30k_b}.


\subsection{Robust Adapter (R-Adapter)}
\label{sec:R-Adapter}
To achieve efficient and robust fine-tuning, we introduce R-Adapter. 
Our method is grounded in the PEFT framework, which freezes the pre-trained model while tuning a small number of additional learnable parameters. However, a na\"ive application of this framework in training can incur a significant bias towards in-distribution data
(refer to Table~\ref{tab:ablation_components}). Drawing inspiration from observations that ensembles enhance generalizability across a wide range of distributions~\cite{izmailov2018averaging, wiseft}, R-Adapter is designed with three novel self-ensembling strategies to enable robust fine-tuning without adding computational load during training and inference.
In the following, we will introduce the design of R-adapter and then describe our three self-ensemble strategies.

\begin{figure*} [!t]
\centering
\includegraphics[width = 0.95\linewidth]{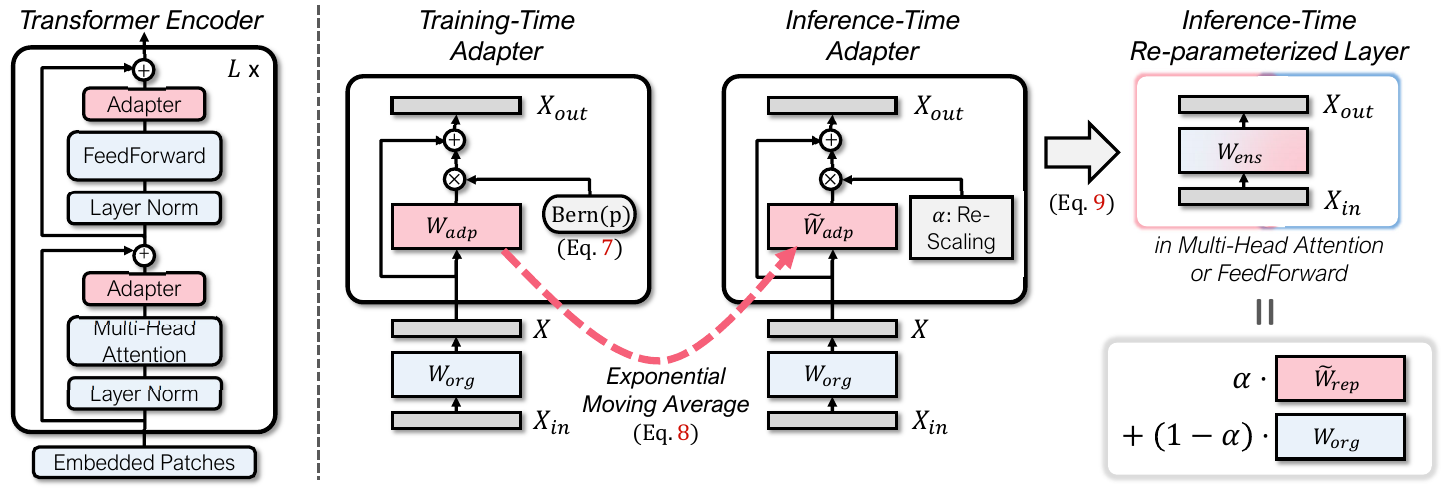}
\caption{An overview of R-Adapter. Each adapter is positioned after MHA and FFN layers. R-Adapter stochastically drops the adapters during training. Also, the weights of the adapters are accumulated using an exponential moving average during the training. At the evaluation, these weights are re-scaled by $\alpha$ and then re-parametrized to be integrated into their prior layers, resulting in a weight-space ensemble between the pre-trained layers and the re-parametrized layer without re-scaling.} 
\label{fig:overview}
\end{figure*}

\vspace{1mm}
\noindent\textbf{Design of R-Adapter.}
\label{subsec:design}
R-Adapter builds upon the adapter-tuning framework where lightweight modules are added to a pre-trained model. Specifically, the adapter modules in R-Adapter adopt the simple version of the Houlsby Adapter~\cite{houlsby2019parameter} 
removing nonlinear layers and bias.
The module is structured as a residual block composed of a weight matrix as follows:
\begin{equation}
    h(X) = XW_{\textrm{adp}} + X,
    \label{eq:Adapter}
\end{equation}
where $X$ means an output of a pre-trained block and $W_{\textrm{adp}} \in \mathbb{R}^{d\times d}$ is the weight matrix of our adapter. For full-shot learning, we maintain a full-rank structure for $W_{\textrm{adp}}$ to preserve sufficient capacity. In the few-shot learning, we can adopt a bottleneck architecture by decomposing $W_{\textrm{adp}}$ into a product of low-rank matrices $BA$, where $B\in \mathbb{R}^{d\times r}$, $A\in \mathbb{R}^{r\times d}$, and the rank $r \ll d$.  This decomposition avoids over-parameterization and significantly reduces the number of parameters and computations. We deploy adapters per Transformer layer in \emph{both image and text encoders}, positioned after MHA and FFN layers, as shown in Fig.~\ref{fig:overview}.

Since our adapters lack nonlinearity in between, we can re-parameterize the adapter to remove extra computation overhead from adapter during inference by integrating it with the closest pre-trained layer~\cite{repadapter}. 
The weights of the pre-trained layer preceding the adapter, denoted by $W_{\textrm{org}}$, is either $W_O$ from MHA (Eq.~\ref{eq:MHA}) or $W_2$ in FFN (Eq.~\ref{eq:FFN}), and the corresponding bias $b_{\textrm{org}}$ is $b_2$ in FFN (Eq.~\ref{eq:FFN}). 
Given the input to pre-trained layers $X_{\textrm{in}}$, the re-parametrization is then conducted by

\begin{gather}
\begin{aligned}
   h(X_\textrm{in}W_\textrm{org} + b_\textrm{org}) &= X_\textrm{in}W_\textrm{org}(W_{\textrm{adp}} + \mathrm{I})  + b_{\textrm{org}}W_{\textrm{adp}} + b_{\textrm{org}} \\
    &= X_\textrm{in}W_\textrm{rep} + b_\textrm{rep},
    \label{eq:rep}
\end{aligned}
\raisetag{12pt}
\end{gather}
where $\mathrm{I}\in\mathbb{R}^{d\times d}$ is the identity matrix, $W_\textrm{rep} = W_\textrm{org}(W_\textrm{adp}+\mathrm{I})$, and $b_\textrm{rep} = b_\textrm{org}(W_\textrm{adp}+\mathrm{I})$. 


\vspace{1mm}
\noindent\textbf{Dynamic Ensemble by Adapter Dropping.}
\label{subsec:stochastic}
To enhance OOD robustness, R-Adapter employs a dynamic ensemble technique through adapter dropping. During only training, adapter modules are randomly deactivated as follows:

\begin{equation}
    \centering
    h(X) = \frac{\gamma}{1-p} \cdot XW_{\textrm{adp}} + X,
    \label{eq:Stochastic}
\end{equation}
where $\gamma$ is an independent variable drawn from $\textrm{Bernoulli}(1-p)$, and $p$ is the drop probability of the adapter dropping.
Unlike dropout~\cite{Srivastava2014} for feature sparsity or drop-path~\cite{larsson2016fractalnet} for model depth reduction, our technique uniquely focuses on randomly disabling adapter layers while consistently supplying pre-trained features.
Adapter dropping is not applied during inference, serving to create an ensemble of subnetworks that vary by the combination of both pre-trained and adapter layers. This strategy enables a dynamic ensemble of multiple models that retain both pre-trained knowledge and fine-tuned knowledge simultaneously and thus boost performance both on ID and OOD data. (see Table~\ref{tab:ablation_components})

\noindent\textbf{Temporal Ensemble by Accumulation.}
\label{subsec:accumulation}
We advance the robustness of the model by incorporating a temporal ensemble strategy through the historical accumulation of adapter weights. The ensemble captures a broader understanding of the feature space by averaging the weights over multiple iterations during training~\cite{izmailov2018averaging, cha2021swad}. The weights of the accumulated adapter $\tilde{W}_\textrm{adp}$ are updated via an exponential moving average:
\begin{equation}
    \tilde{W}_\textrm{adp} \leftarrow m \cdot \tilde{W}_\textrm{adp} + (1-m) \cdot {W}_\textrm{adp}, 
    \label{eq:Accumulation}
\end{equation}
where $m \in [0, 1]$ is the coefficient that controls the momentum update rate. This procedure is notably \emph{memory-efficient} since only the parameters of adapters are momentum updated, not the parameters of the entire model. In inference time, we utilize the accumulated weights $\tilde{W}_\textrm{adp}$ for Eq.~\ref{eq:rep}, thereby produces re-parameterized weight $\tilde{W}_\textrm{rep}$ and bias $\tilde{b}_\textrm{rep}$.

\noindent\textbf{Weight-space Ensemble by Re-scaling.}
\label{subsec:rescale}
Finally, we introduce a strategy that establishes a weight-space ensemble between the pre-trained and fine-tuned layers through re-scaling with re-parameterization. 
The conventional weight-space ensemble (WiSE-FT)~\cite{wiseft} linearly interpolates between the weights of the original pre-trained parameters and the fine-tuned parameters, thus requiring storing both separate models.
In contrast, we evolve this concept by employing the re-parameterized weights $\tilde{W}_\textrm{rep}$ as the weights of a fine-tuned layer. We streamline the weight-space ensemble within a single model to be implemented simply by re-scaling the weights of the adapter and re-parameterizing them at inference. 
This process is expressed as follows:
\begin{gather}
\begin{aligned}
   \underbrace{\alpha \tilde{W}_\textrm{rep} + (1-\alpha) W_\textrm{org}}_\texttt{{\makebox[0pt]{\scalebox{0.95}{Weight-space Ensemble}}}} &= \alpha W_\textrm{org}\tilde{W}_\textrm{adp}  + \alpha W_\textrm{org} + (1-\alpha) W_\textrm{org} \\[-17pt]
   &= \underbrace{W_\textrm{org}(\overbrace{\alpha \tilde{W}_\textrm{adp}}^\texttt{{\makebox[0pt]{\scalebox{0.95}{Re-scaling}}}} \;+\; \mathrm{I}) = W_\textrm{ens}}_\texttt{\makebox[0pt]{\scalebox{0.95}{Re-parametrization}}}, \\[-21pt]
    \label{eq:rescale}
\end{aligned}-
\raisetag{17pt}
\end{gather}
where $W_\textrm{ens}$ denotes the ensembled weights, and $\alpha$ is a re-scaling coefficient.
The coefficient $\alpha$ serves as an interpolation factor, adjusting the balance between the original pre-trained weights 
$W_\textrm{org}$ and the adjusted weights of the fine-tuned layer. 
This technique not only improves accuracy under distribution shifts but also maintains high performance on the ID data. 
Crucially, unlike WiSE-FT, our method does not require maintaining two separate full models in storage, thus 
facilitating weight-space ensemble \emph{more storage-efficiently}.


\subsection{MPM-NCE Loss for Downstream Task}
\label{sec:MPM_NCE}

To enhance learning for downstream tasks, it is crucial to use loss functions that align closely with the characteristics of the tasks. Vision-language tasks often involve multiple correspondences between modalities. For instance, in classification tasks, using different text templates for the same class can result in multiple text descriptions matching a single image, and naturally the reverse is true as well. This situation also occurs in cross-modal retrieval tasks with images and captions.
When adapting zero-shot models to new tasks, a common approach is to use the InfoNCE loss used for pre-training. However, this loss is not ideal for tasks where multiple positive samples exist, as it considers a single positive pair. Moreover, InfoNCE learns the ordering between positive and negative samples, which may not lead to sufficiently discriminative features for downstream tasks.

To address these limitations, we propose MPM-NCE Loss, designed to accommodate the multi-positive nature of these tasks while enhancing the discriminative power of the learned embeddings. This loss function has two pivotal improvements. First, we use soft labels that assign equal probability to multiple positive pairs. 
The formulation of the soft label is given as follows:
\begin{equation}
    \tilde{y}_{ij} = \frac{(1-\epsilon)\cdot y_{ij}}{|P(i)|}
    + \frac{\epsilon \cdot (1-y_{ij})}{B-|P(i)|} \in [0,1],
    \label{eq:soft_label} 
\end{equation}
where $y_{ij} \in \{0,1\}$  indicates the positive relation between samples $i$ and $j$, $P(i)$ is the set of positive samples of sample $i$ including itself and $\epsilon$ is a label smoothing noise~\cite{szegedy2016rethinking}.
This soft label ensures the correct alignment of multiple image-text pairs in downstream tasks. Additionally, the soft labels can include $\epsilon$, reducing overfitting risks by introducing a minor perturbation to the labels. 

The second improvement is the addition of a margin $\delta$ applied to negative pairs. This margin enhances the
discrimination of learned features by ensuring that negative pairs are not only distinct but separated by a certain threshold.  Incorporating these improvements, our MPM-NCE is formulated as follows:

\vspace{-3mm}
\begin{equation}
\fontsize{9.6}{10}\selectfont
    \mathcal{L}(\mathcal{B}) = -\sum_{i,j=1}^{B}\Bigg(\tilde{y}_{ij}\log\frac{e^{(f_i \cdot g_j+\delta_{ij})/\tau }}{\sum_{k=1}^{B}e^{(f_i \cdot g_k + \delta_{ik})/\tau}}
    +\tilde{y}_{ji}\log\frac{e^{ (f_j \cdot g_i+\delta_{ji})/\tau}}{\sum_{k=1}^{B}e^{(f_k \cdot g_i + \delta_{ki})/\tau}}\Bigg),
    \label{eq:MPM_NCE}
\end{equation}
where the temperature $\tau$ is set to a constant value of 0.01, and $\delta_{ij}$ is 0 for positive relations and $\delta$ for the rest.
Consequently, MPM-NCE loss encourages the model to correctly align multiple image-text pairs and learn discriminative features, leading to notable improvements in performance under ID and OOD.


\section{Experiments}
\label{sec:experiments}

We first demonstrate the robustness of R-Adapter against natural distribution shifts for image classification and its efficiency (Sec.~\ref{sec:classification}). We then analyze the effectiveness of proposed components in R-Adapter and MPM-NCE loss, including ensemble techniques and loss, compared to existing approaches and also conduct an ablation study on hyperparameters. Furthermore, we validate the versatility of R-Adapter by extending it to broader tasks such as few-shot classification (Sec.~\ref{sec:fewshot}), cross-modal retrieval (Sec.~\ref{sec:cross_retrieval}), open-vocabulary segmentation (Sec.~\ref{sec:ovseg}), and base-to-novel generalization (in Appendix).

\subsection{Datasets}\label{sec:datasets}


\noindent{\textbf{Image Classification.}}
We use ImageNet (IN)~\cite{Imagenet} as the ID dataset for fine-tuning; we evaluate the robustness of the models on five standard OOD datasets with different distribution shifts, following prior work~\cite{wiseft,flyp,maskfill,clip,lpft}: ImageNetV2 (IN-V2)~\cite{recht2019imagenet}, 
ImageNet-R (IN-R)~\cite{hendrycks2021many}, ImageNet-Sketch (IN-Sketch)~\cite{wang2019learning}, ObjectNet~\cite{barbu2019objectnet}, and ImageNet-A (IN-A)~\cite{hendrycks2021natural}. Note that these datasets except ObjectNet are also used in a few-shot setting following previous work~\cite{cocoop, coop, repadapter,clipood}.

\noindent{\textbf{Cross-Modal Retrieval.}}
We utilize two standard benchmarks for image-text cross-modal retrieval, COCO~\cite{Mscoco} as ID and Flickr30K~\cite{Flickr30k_b} as OOD. For these two datasets, each image is associated with the corresponding five captions.

\noindent{\textbf{Open-Vocabulary Segmentation.}}
Following our baseline method~\cite{ovseg}, we train a CLIP model on the COCO Captions dataset~\cite{chen2015microsoft} and test it on several OOD benchmarks: ADE20K~\cite{zhou2019semantic} (A-150 and A-847 category versions), Pascal Context~\cite{mottaghi2014role} (PC-59 and PC-459 category versions), and Pascal VOC~\cite{Pascalvoc}.

\subsection{Implementation Details}\label{sec:details}
\noindent{\textbf{Network Architectures.}}
We adopt the pre-trained CLIP models from OpenAI~\cite{clip} with four different sizes of image encoder, {ViT-B/32}, {ViT-B/16}, {ViT-L/14}, and {ViT-L/14@336px~\cite{ViT}}.

\noindent{\textbf{Network Optimization.}}
Our model is trained using AdamW without weight decay for 10 epochs, except for open vocabulary segmentation which is trained for 5 epochs following previous work~\cite{ovseg}.
The initial learning rate is set to $\expnum{5}{4}$, using a cosine scheduling with 500 warm-up steps.
We closely follow the settings in~\cite{flyp} for full-shot classification,~\cite{clipood} for few-shot classification, and~\cite{ovseg} for open-vocabulary segmentation. More details are in the appendix.


\noindent{\textbf{Hyperparameters.}}
The drop probability $p$ is set to 0.2.
The momentum update rate $m$ in Eq.~\ref{eq:Accumulation} is set to 0.999.
The margin $\delta$ in Eq.~\ref{eq:MPM_NCE} is 0.05.
For classification tasks, following the WiSE-FT~\cite{wiseft}, we use the re-scaling coefficient $\alpha$ in Eq.~\ref{eq:rescale} of 0.5. 
For cross-modal retrieval and open vocabulary segmentation tasks, we set $\alpha$ to its optimal values of 0.8 and 0.4, respectively.
We set the smoothing coefficient $\epsilon$ in Eq.~\ref{eq:soft_label} to 0.05 for classification, and 0 for other tasks.


\subsection{ImageNet Classification Under Distribution Shifts} 

\setlength{\floatsep}{10pt plus 0pt minus 5pt}
\setlength{\textfloatsep}{10pt plus 0pt minus 5pt}
\begin{table*}[!t]
\caption{Top-1 accuracy of models with different robust fine-tuning on ImageNet (ID) and OOD datasets. ``OOD avg'' is the average accuracy across the five OOD datasets. 
Entries in \textcolor{grn}{green} indicate fewer parameters than full fine-tuning, and \textcolor{red}{red} use more.
}
\setlength{\tabcolsep}{2pt}
\fontsize{6.8}{8.7}\selectfont
\centering
\scalebox{1.0}
{\begin{tabularx}{1.0\textwidth}
    {
      p{0.182\textwidth}
      >{\centering\arraybackslash}p{0.15\textwidth}
      >{\centering\arraybackslash}p{0.063\textwidth}
      >{\centering\arraybackslash}p{0.11\textwidth} |
      >{\centering\arraybackslash}p{0.063\textwidth}
      >{\centering\arraybackslash}p{0.063\textwidth}
      >{\centering\arraybackslash}p{0.1\textwidth}
      >{\centering\arraybackslash}p{0.1\textwidth} 
      >{\centering\arraybackslash}p{0.063\textwidth}
      }
     \toprule
    {\multirow{1}{*}[-4mm]{\textbf{Methods}}}& \multirow{1}{*}[-0.5mm]{\textbf{Trainable}}&
    \multicolumn{1}{c}{\textbf{ID}} & \multicolumn{6}{c}{\textbf{Out-Of-Distribution~(OOD)}} \\ [-0.3ex]  \cmidrule(lr){3-3}  
    \cmidrule(lr){4-9} &\multirow{1}{*}[0.5mm]{\textbf{ Params (M)}} & IN & {OOD avg} & IN-V2 & IN-R&  {IN-Sketch} &{ObjectNet}& IN-A  \\  [-0.4ex] \midrule
    \multicolumn{9}{l}{\textit{\textbf{CLIP ViT-B/32}}} \\ [-0.3ex]\midrule
     Zero-Shot~\cite{clip} & \xmark& 63.4 &  48.7  &55.9 & 69.3 & 42.3 & 44.5 & 31.4 \\
     Fine-Tuning~(FT) & {88.4} & 75.9 & 44.2 & 64.7 & 57.0 & 39.8 & 39.5 & 20.0  \\
     WiSE-FT~\cite{wiseft} & {88.4}& 76.6 &52.4 & 66.6 & \underline{70.2} & 47.1 &46.3 &31.9   \\
     {Uniform Soup~\cite{modelsoup}} &\textcolor{red}{6364.8}  & \textbf{80.0} &51.6&\textbf{68.6} & 66.6 &\underline{47.7} &46.1 &29.2  \\
     Mask-Fill~\cite{maskfill} &{88.4}& 77.5&\underline{53.1} &67.1 &69.7 &46.9 &\underline{48.0} &\underline{33.8}  \\ 
     \ccol  Ours &\ccol  \textcolor{grn}{{20.5}} & \underline{77.7} \ccol & \ccol \textbf{54.3}&\ccol  \underline{67.7} &\ccol \textbf{70.8} &  \ccol  \textbf{47.8}   & \ccol \textbf{49.7} & \ccol \textbf{35.6}   \\\midrule
    \multicolumn{9}{l}{\textit{\textbf{CLIP ViT-B/16}}} \\ [-0.4ex] \midrule
     Zero-Shot~\cite{clip} & \xmark &  68.3 &58.4 &61.9 &77.6 &48.3 &54.0 &50.1   \\
     Fine-Tuning~(FT) & {86.7}& 80.7 & 52.8 &70.4 &64.0 &45.1 &49.1 &35.2  \\
     LP-FT~\cite{lpft} & {86.7}&  81.7 &60.3 &72.1 &73.5 &50.3 &58.2 &47.6 \\    
     WiSE-FT~\cite{wiseft} & {86.7}& 81.7&63.0 &72.8 &\underline{78.7} &\textbf{53.9} &57.3 &52.2   \\
     FLYP~\cite{flyp} & \textcolor{red}{149.6}& \underline{82.6} &60.2 &73.0 &71.4 &48.1 &58.7 & 49.6   \\
     WiSE-FLYP ~\cite{flyp}&  \textcolor{red}{149.6} & \textbf{82.9} &63.1 &\underline{73.5} &76.0 &53.0 &\textbf{60.8} &52.3  \\
     Mask-Fill~\cite{maskfill}& {86.7}& 82.4&\underline{63.3}& 73.4& 78.1 &53.4 &57.9& \underline{53.5}  \\
     \ccol  Ours &\ccol  \textcolor{grn}{20.5}& {82.0} \ccol & \ccol \textbf{64.8} &\ccol  \textbf{73.6} &\ccol \textbf{79.1} & \ccol \textbf{53.9} & \ccol \underline{59.7}  & \ccol \textbf{57.5}   \\ \midrule
    \multicolumn{9}{l}{\textit{\textbf{CLIP ViT-L/14@336px}}} \\ [-0.4ex] \midrule
     WiSE-FT~\cite{wiseft}& 305.1&\textbf{86.8}& \underline{76.9}& \underline{79.5} &\underline{89.4} &\textbf{64.7} &\underline{71.1} &\underline{79.9} \\
     \ccol  Ours &\ccol  \textcolor{grn}{64.5}& \textbf{86.8} \ccol & \ccol \textbf{78.9} &\ccol  \textbf{79.6} &\ccol \textbf{89.9} & \ccol \underline{64.1} & \ccol \textbf{73.3}  & \ccol \textbf{82.4}   \\
\bottomrule 
\end{tabularx}
}
\label{tab:comparison_finetuned}
\end{table*}

\subsubsection{Main Results.}\label{sec:classification}

We compare our method with zero-shot, conventional fine-tuning approach, and previous robust fine-tuning methods, WiSE-FT~\cite{wiseft}, LP-FT~\cite{lpft}, Model Soup~\cite{modelsoup}, FLYP~\cite{flyp}, and Mask-Fill~\cite{maskfill} on the in-distribution (ID) dataset and five out-of-distribution (OOD) datasets.
We report the performance of WiSE-FT with the default mixing coefficient of 0.5. We take the \textit{uniform soup} as the default method of~\cite{modelsoup}.
The results and the number of trainable parameters are summarized in Table~\ref{tab:comparison_finetuned}.
Specifically, our method improves the previous state of the art by a significant margin as 1.2\%p and 1.5\%p in terms of OOD avg. with CLIP ViT-B/32 and CLIP ViT-B/16, respectively; even though our method only requires much less tunable parameters (20.5M) than the others ($>$80M).
Moreover, our method scales efficiently to the CLIP ViT-L/14@336px model, showing a notable 2\%p improvement in OOD performance over WiSE-FT. 
While the Uniform Soup achieves superior results on IN and IN-V2, it involves a complex ensemble of fine-tuned models, leading to increased computational and resource demands. In contrast, our method offers a cost-efficient approach to enhancing robustness, as evidenced by the pronounced gains observed in the most distribution-shifted datasets, IN-A and IN-R.

\setlength{\floatsep}{10pt plus 0pt minus 5pt}
\setlength{\textfloatsep}{13pt plus 0pt minus 5pt}
\begin{table}[!t]
\caption{Ablation study on key components of our method and comparison with the other adapter-tuning methods using full-rank structure. The experiments are performed on the ImageNet classification with ViT-B/32. 
The last row (\textbf{E10}) corresponds to our default configuration. DO: Dropout in Adapters. DP: Drop-path in pre-trained layers. AD: Adapter Dropping. AC: Accumulation. RS: Re-scaling. LS: Label Smoothing.}
\setlength{\tabcolsep}{1pt}
\fontsize{6.5}{7.1}\selectfont
\centering
\scalebox{1.0}
{\begin{tabularx}{1.0\columnwidth}
{
  >{\centering\arraybackslash}p{0.045\textwidth} | 
  >{\centering\arraybackslash}p{0.175\textwidth} |
  >{\centering\arraybackslash}p{0.045\textwidth}
  >{\centering\arraybackslash}p{0.045\textwidth} 
  >{\centering\arraybackslash}p{0.045\textwidth}
  >{\centering\arraybackslash}p{0.045\textwidth}
  >{\centering\arraybackslash}p{0.045\textwidth} 
  >{\centering\arraybackslash}p{0.1\textwidth} 
  >{\centering\arraybackslash}p{0.12\textwidth} 
  >{\centering\arraybackslash}p{0.045\textwidth}
  >{\centering\arraybackslash}X 
  >{\centering\arraybackslash}X 
  }
     \toprule
    \multirow{1}{*}[-0.6mm]{Exp}&\multirow{1}{*}[-0.6mm]{\textbf{Adapter Design}}
    & \multicolumn{5}{c}{\textbf{Regularization}} & \multicolumn{3}{c}{\textbf{Loss}} & \multicolumn{2}{c}{\textbf{Accuracy}} \\ \cmidrule(lr){3-7} \cmidrule(lr){8-10} \cmidrule(lr){11-12}    
    \multirow{1}{*}[0.6mm]{No.}&\multirow{1}{*}[0.6mm]{{(w/ Full-Rank)}}& {{DO}} & {{DP}} &{\textbf{AD}}&{\textbf{AC}}& {\textbf{RS}}  & {{InfoNCE}}& {\textbf{MPM-NCE}}&  \textbf{LS} & {{ID}}& {{OOD avg}}\\  \midrule
    B1&AdaptFormer~\cite{adaptformer} & \cmark& \textcolor{grey}{\xmark} & \textcolor{grey}{\xmark}&\textcolor{grey}{\xmark} &\textcolor{grey}{\xmark}& \cmark & \textcolor{grey}{\xmark}& \textcolor{grey}{\xmark}& 77.2& 48.5 \\
    B2&RepAdapter~\cite{repadapter} & \cmark& \cmark& \textcolor{grey}{\xmark}&\textcolor{grey}{\xmark} &\textcolor{grey}{\xmark}& \cmark & \textcolor{grey}{\xmark}& \textcolor{grey}{\xmark}& 77.2& 48.3 \\ \midrule
    E0&& \textcolor{grey}{\xmark} & \textcolor{grey}{\xmark} &\textcolor{grey}{\xmark}&\textcolor{grey}{\xmark} &\textcolor{grey}{\xmark}& \cmark & \textcolor{grey}{\xmark}& \textcolor{grey}{\xmark}& 77.5 ($\uparrow$ 0.0)& 47.7 ($\uparrow$ 0.0) \\ 
    E1&& \cmark& \textcolor{grey}{\xmark} & \textcolor{grey}{\xmark}&\textcolor{grey}{\xmark} &\textcolor{grey}{\xmark}& \cmark & \textcolor{grey}{\xmark}& \textcolor{grey}{\xmark}& 77.6 \textcolor{grn}{($\uparrow$ 0.1)}& 48.7 \textcolor{grn}{($\uparrow$ 1.1)}\\
    E2&& \textcolor{grey}{\xmark} & \cmark& \textcolor{grey}{\xmark}&\textcolor{grey}{\xmark} &\textcolor{grey}{\xmark}& \cmark & \textcolor{grey}{\xmark}& \textcolor{grey}{\xmark}& 77.4 \textcolor{red}{($\downarrow$ 0.1)}& 47.9 \textcolor{grn}{($\uparrow$ 0.2)}\\ 
    E3&& \textcolor{grey}{\xmark} & \textcolor{grey}{\xmark} & \cmark&\textcolor{grey}{\xmark} &\textcolor{grey}{\xmark}& \cmark & \textcolor{grey}{\xmark}& \textcolor{grey}{\xmark}& 77.8 \textcolor{grn}{($\uparrow$ 0.3)}& 49.6 \textcolor{grn}{($\uparrow$ 1.9)}\\
    E4&\multirow{1}{*}[-1mm]{R-Adapter}&\textcolor{grey}{\xmark} & \textcolor{grey}{\xmark} & \textcolor{grey}{\xmark}&\cmark&\textcolor{grey}{\xmark}& \cmark & \textcolor{grey}{\xmark}& \textcolor{grey}{\xmark}& 77.4 \textcolor{red}{($\downarrow$ 0.1)}& 47.8 \textcolor{grn}{($\uparrow$ 0.1)} \\
    E5&\multirow{1}{*}[-1mm]{(Ours)} &\textcolor{grey}{\xmark} & \textcolor{grey}{\xmark} & \textcolor{grey}{\xmark}&\textcolor{grey}{\xmark} &\cmark& \cmark & \textcolor{grey}{\xmark}& \textcolor{grey}{\xmark}& 76.5 \textcolor{red}{($\downarrow$ 1.0)}& 53.5 \textcolor{grn}{($\uparrow$ 5.8)} \\
    E6&&\textcolor{grey}{\xmark} & \textcolor{grey}{\xmark} & \cmark&\cmark&\textcolor{grey}{\xmark}& \cmark & \textcolor{grey}{\xmark}& \textcolor{grey}{\xmark}& 77.9 \textcolor{grn}{($\uparrow$ 0.4)}&49.9 \textcolor{grn}{($\uparrow$ 2.2)} \\ 
    E7&&\textcolor{grey}{\xmark} & \textcolor{grey}{\xmark} & \cmark&\cmark&\cmark& \cmark & \textcolor{grey}{\xmark}& \textcolor{grey}{\xmark}&  76.6 \textcolor{red}{($\downarrow$ 0.9)}& 53.7 \textcolor{grn}{($\uparrow$ 6.0)}\\
    E8&&\textcolor{grey}{\xmark} & \textcolor{grey}{\xmark} & \cmark&\cmark&\cmark&  \cmark& \textcolor{grey}{\xmark} & \cmark & 76.9 \textcolor{red}{($\downarrow$ 0.6)}& 54.0 \textcolor{grn}{($\uparrow$ 6.3)} \\ 
    E9&&\textcolor{grey}{\xmark} & \textcolor{grey}{\xmark} & \cmark&\cmark&\cmark& \textcolor{grey}{\xmark} & \cmark& \textcolor{grey}{\xmark}&  77.5 ($\uparrow$ 0.0)& 53.9 \textcolor{grn}{($\uparrow$ 6.2)}\\ 
    \textbf{E10}&& \textcolor{grey}{\xmark} & \textcolor{grey}{\xmark} & \cmark& \cmark& \cmark&  \textcolor{grey}{\xmark} & \cmark& \cmark&  77.7 \textcolor{grn}{($\uparrow$ 0.2)}& 54.3 \textcolor{grn}{($\uparrow$ 6.6)} \\
\bottomrule 
\end{tabularx}
}
\label{tab:ablation_components}
\end{table}

\noindent{\textbf{Effectiveness of Key Components.}}
In our ablation study, we evaluate the impact of key components and compare our method with AdaptFormer and RepAdapter, both trained with the FLYP scheme, as shown in Table~\ref{tab:ablation_components}. Despite using regularization techniques like dropout (DO) and drop-path (DP), these methods perform poorly in out-of-distribution (OOD) settings, revealing the limitations of na\"ively combining PEFT with robust fine-tuning.
Our base R-Adapter model (E0) also falls short in OOD accuracy. However, using Adapter Dropping (AD) improves OOD accuracy by 1.9\% and in-distribution (ID) accuracy by 0.3\% (E1, E2, and E3). Accumulation (AC) and Re-scaling (RS) are crucial for OOD robustness (E4 and E5), with RS boosting OOD performance by 5.8\% despite a slight reduction in ID performance. Combining our regularization techniques mitigates this reduction and further enhances OOD accuracy (E6 and E7).
MPM-NCE outperforms InfoNCE in both ID and OOD settings by 0.9\% and 0.2\%, respectively (E7 and E9). While label smoothing (LS) with InfoNCE can reduce ID performance due to semantic misalignments, MPM-NCE with LS improves both ID and OOD performance by maintaining accurate alignment and providing additional regularization (E10). 
Our default model, the R-Adapter trained with MPM-NCE loss, significantly advances ID performance and OOD robustness over existing adapter techniques (B1, B2, and \textbf{E10}).

\begin{wrapfigure}{r}{0.491\textwidth}
\vspace{-10mm}
\begin{center}
\includegraphics[width = 0.491\textwidth]{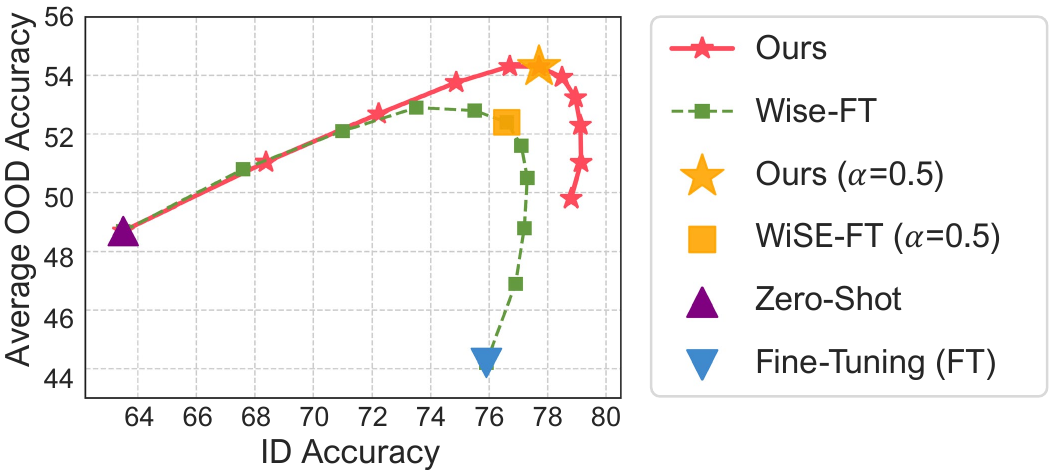}
\end{center}
\vspace{-5mm}
\caption{
Performance of our method varying re-scaling coefficient $\alpha $ against WiSE-FT.} 
\label{fig:rescaling}
\vspace{-5mm}
\end{wrapfigure}

\setlength{\floatsep}{10pt plus 0pt minus 5pt}
\setlength{\textfloatsep}{12pt plus 0pt minus 5pt}
\begin{table}[!t]
\caption{Ablation study on hyperparameters on the ImageNet classification task with ViT-B/32. The last column shows the average accuracy across the five OOD datasets. \colorbox{grey}{{gray}} corresponds to our default setting. ``w/ SP'' indicates the considering single positive without soft labels as InfoNCE, but employing a margin $\delta$ of 0.05.}
\vspace{-3.5mm}
    \centering
    \fontsize{5.8}{7}\selectfont        
        \scalebox{1.0}{
 	\begin{subtable}[t]{0.3\textwidth}
		\caption{Rank of Adapter} 
        \vspace{-3mm}
        \setlength{\tabcolsep}{2pt}
		\centering
		{\begin{tabular}{c|c|c|c}
        \toprule
		Rank & \#Params & ID& OOD\\
		\midrule
            4 & 0.25M& 72.5 & 51.7  \\
            8 & 0.49M& 73.4&52.4  \\
            16 & 0.98M& 74.5& 52.5 \\
		128 & 7.84M& 76.7 & 53.7 \\
            \ccol Full& \ccol 20.45M& \ccol \textbf{77.7}& \ccol \textbf{54.3} \\
        \bottomrule
		\end{tabular}}
		\label{tab:rank}
	\end{subtable}
    \hfill
	\begin{subtable}[t]{0.23\textwidth} 
		\caption{Loss Variations} 
  \vspace{-3mm}
        \setlength{\tabcolsep}{3pt}
		\centering
		{\begin{tabular}{c|c|c}
        \toprule
		Loss & ID& OOD\\
		\midrule
		$\delta=0$ &  77.1&  54.0\\
        $\delta=0.02$ & 77.5& 54.3\\
        \ccol $\delta=0.05$ &\ccol {77.7}& \ccol \textbf{54.3} \\
        $\delta=0.1$ & \textbf{77.8}& 53.8 \\
        w/ SP  & 77.2& 47.0 \\
        \bottomrule
		\end{tabular}}
		\label{tab:margin}
	\end{subtable}
    \hfill
	\begin{subtable}[t]{0.21\textwidth}
 		\caption{$p$ in Eq.~\ref{eq:Stochastic}}
            \vspace{-3mm}
		\label{tab:stochastic}
        \setlength{\tabcolsep}{4pt}
        \centering
		{\begin{tabular}{c|c|c}
        \toprule
		$p$ & ID& OOD\\
		\midrule
		 0 &  77.6&  53.3\\
        0.1 & \textbf{77.9}& 54.0\\
        \ccol {0.2} & \ccol 77.7& \ccol 54.3\\
        0.3 & 77.6& \textbf{54.4}\\
        0.5 &  77.1& 54.2\\
        \bottomrule
		\end{tabular}}
	\end{subtable}
	\hfill
	\begin{subtable}[t]{0.21\textwidth}
        \caption{$m$ in Eq.~\ref{eq:Accumulation}}
        \vspace{-3mm}
        \setlength{\tabcolsep}{4pt}
        \centering
		{\begin{tabular}{c|c|c}
        \toprule
		$m$ & ID& OOD\\
		\midrule
		 0 &  \textbf{77.8}&  54.0\\
        0.9 & \textbf{77.8}& \textbf{54.3}\\
        0.99 & \textbf{77.8}& \textbf{54.3}\\
        \ccol {0.999} & \ccol 77.7& \ccol \textbf{54.3}\\
        0.9999 & 77.0& \textbf{54.3}\\
        \bottomrule
		\end{tabular}}		
		\label{tab:ema}
	\end{subtable}
    \hfill
    }
    \label{tab:ablation_hyperparams}
\end{table}

\noindent{\textbf{Effect of Hyperparameters.}} 
We investigate the effects of the rank of the adapter module $r$ and various hyperparameters, including the drop probability $p$ in Eq.~\ref{eq:Stochastic}, the momentum update rate $m$ in Eq.~\ref{eq:Accumulation} and the margin of our loss $\delta$ in Eq.~\ref{eq:MPM_NCE}. Table ~\ref{tab:rank} reveals that increasing the rank of the adapter enhances performance, due to improved model capacity. This result aligns with findings in \cite{chen2022revisiting} that more parameters yield better results in data-rich environments.  
Table~\ref{tab:ablation_hyperparams} shows gradual performance gains with a margin $\delta$ up to 0.05, but using a margin with a single positive reduces OOD performance.
As shown in Table~\ref{tab:stochastic} and~\ref{tab:ema}, each hyperparameter brings performance improvement compared to when it is set to 0, regardless of specific values.
Fig.~\ref{fig:rescaling} shows the impact of varying the re-scaling coefficient $\alpha$ in Eq.~\ref{eq:rescale}. Compared to WiSE-FT~\cite{wiseft}, our method shows less sensitivity to changes in $\alpha$, maintaining superior performance across various settings.

\begin{table}[!t]
\caption{Top-1 accuracy for adapting CLIP to 16-shot ImageNet classification on ID and OOD datasets. {OOD avg} is the average accuracy across the four OOD datasets. ``$r$-Rank'' denotes our models with adapters employing low-rank decomposition while ``Full-Rank'' is no decomposition. All methods adopt CLIP ViT-B/16 as the backbone.}
\label{tab:fewshot}
\fontsize{6.6}{8.6}\selectfont
\centering
\scalebox{0.98}{
\begin{tabularx}{1.0\textwidth}{%
    p{0.19\textwidth} 
    >{\centering\arraybackslash}p{0.16\textwidth} 
    >{\centering\arraybackslash}p{0.09\textwidth} 
    >{\centering\arraybackslash}p{0.12\textwidth} |
    >{\centering\arraybackslash}p{0.09\textwidth} 
    >{\centering\arraybackslash}p{0.09\textwidth} 
    >{\centering\arraybackslash}p{0.1\textwidth} 
    >{\centering\arraybackslash}p{0.09\textwidth}}
    \toprule
    \multirow{2}{*}[-3mm]{\textbf{Methods}} & {\multirow{1}{*}[-0.5mm]{\textbf{Trainable}}} & \multicolumn{1}{c}{\textbf{ID}} & \multicolumn{5}{c}{\textbf{Out-Of-Distribution~(OOD)}} \\ 
    \cmidrule(lr){3-3} \cmidrule(lr){4-8}
    & \multirow{1}{*}[0.5mm]{{\textbf{Params (M)}}} & {IN} & {OOD avg} & {IN-V2} & {IN-R} & {IN-Sketch} & {IN-A} \\ [-0.3ex] 
    \midrule
    Zero-Shot~\cite{clip} & \xmark & 68.3 & 58.4 & 61.9 & 77.6 & 48.3 & 50.1 \\
    CoOp~\cite{coop} & $>$ 0.01 & 71.5 & 59.3 & 64.2 & 75.2 & 48.0 & 49.7 \\
    CoCoOp~\cite{cocoop} & 0.03 & 71.0 & 59.9 & 64.1 & 76.2 & 48.8 & 50.6 \\
    RepAdapter-T~\cite{repadapter} & 0.27 & 71.9 & 60.4 & 64.8 & 76.5 & 49.3 & 51.1 \\
    CLIPood~\cite{clipood} & 86.70 & 71.6 & 60.4 & 64.9 & 77.2 & 49.3 & 50.4 \\
    \ccol Ours (1-Rank) & \ccol 0.06 & \ccol 71.7 & \ccol \underline{61.6} & \ccol 65.3 & \ccol \underline{78.6} & \ccol \underline{50.3} & \ccol 52.3 \\
    \ccol Ours (4-Rank) & \ccol 0.25 & \ccol 72.0 & \ccol \underline{61.6} & \ccol 65.1 & \ccol \underline{78.6} & \ccol 50.0 & \ccol \textbf{52.6} \\
    \ccol Ours (8-Rank) & \ccol 0.49 & \ccol \underline{72.4} & \ccol \underline{61.6} & \ccol \underline{65.7} & \ccol \underline{78.6} & \ccol 49.8 & \ccol \underline{52.4} \\
    \ccol Ours (Full-Rank) & \ccol 20.45 & \ccol \textbf{73.9} & \ccol \textbf{62.4} & \ccol \textbf{67.0} & \ccol \textbf{79.1} & \ccol \textbf{51.2} & \ccol 52.3 \\
    \bottomrule
\end{tabularx}}
\end{table}
\subsection{Few-Shot ImageNet Classification}\label{sec:fewshot}
We investigate the robustness of our model when training images are limited, focusing on 16-shot few-shot classification on both ID and OOD datasets. We compare our model with the existing PEFT methods~\cite{coop, cocoop, repadapter} and robust fine-tuning techniques~\cite{clipood}.
As shown in Table~\ref{tab:fewshot}, full-rank R-adapter outperforms the state of the art~\cite{clipood} on all datasets, despite requiring four times fewer trainable parameters.
Furthermore, our model with a rank-1 adapter surpasses CoOp and CoCoOp by 2.3\% and 1.7\% in average OOD top-1 accuracy, with a similar number of tunable parameters. This demonstrates that our method maintains strong generalization on OOD datasets even with extremely minimal parameters.

\setlength{\floatsep}{10pt plus 0pt minus 5pt}
\setlength{\textfloatsep}{13pt plus 0pt minus 5pt}
\begin{table*}[!t]
\caption{
Cross-modal retrieval performance on the COCO (5K test set) and Flickr30K datasets in Recall at K~(R@K). $B$ and $L$ denote the use of 12-layer and 24-layer transformer encoders, respectively. FLYP$_L$ training has failed due to memory constraints.}
\fontsize{6.6}{8.8}\selectfont
\centering
\scalebox{0.98}
{
\begin{tabularx}{1.0\textwidth}
    {
      p{0.192\textwidth}
      >{\centering\arraybackslash}p{0.145\textwidth}
      >{\centering\arraybackslash}X
      >{\centering\arraybackslash}X
      >{\centering\arraybackslash}X
      >{\centering\arraybackslash}X
      >{\centering\arraybackslash}X
      >{\centering\arraybackslash}X
      >{\centering\arraybackslash}X
      >{\centering\arraybackslash}X
      >{\centering\arraybackslash}X
      >{\centering\arraybackslash}X
      >{\centering\arraybackslash}X
      >{\centering\arraybackslash}X
      }
     \toprule
    {\multirow{1}{*}[-7mm]{\textbf{Methods}}}& {\multirow{1}{*}[-2.6mm]{\textbf{Training}}} &
    \multicolumn{6}{c}{\textbf{COCO}} & \multicolumn{6}{c}{\textbf{Flickr30K}} \\ [-0.3ex] 
    \cmidrule(lr){3-8} \cmidrule(lr){9-14}  & &  \multicolumn{3}{c}{\textbf{Text-to-Img}} & \multicolumn{3}{c}{\textbf{Img-to-Text}} & \multicolumn{3}{c}{\textbf{Text-to-Img}} & \multicolumn{3}{c}{\textbf{Img-to-Text}}  \\ [-0.3ex]
    \cmidrule(lr){3-5} \cmidrule(lr){6-8} \cmidrule(lr){9-11} \cmidrule(lr){12-14}  & {\multirow{1}{*}[2.6mm]{\textbf{Dataset}}}& \scalebox{0.94}{R@1} &  \scalebox{0.94}{R@5} & \scalebox{0.94}{R@10} & \scalebox{0.94}{R@1} & \scalebox{0.94}{R@5} & \scalebox{0.94}{R@10} & \scalebox{0.94}{R@1} & \scalebox{0.94}{R@5} & \scalebox{0.94}{R@10} & \scalebox{0.94}{R@1} & \scalebox{0.94}{R@5} & \scalebox{0.94}{R@10} \\ [-0.3ex] \midrule

    {Unicoder-VL$_B$~\cite{li2020unicoder}} & \scalebox{0.88}{Same as Test}& 46.7& 76.0& 85.3 & 62.3 &87.1 &92.8 & 71.5 &90.9 &94.9 & 86.2 &96.3  &99.0  \\     
    Uniter$_L$~\cite{chen2020uniter} & \scalebox{0.98}{Same as Test} & 52.9 &79.9 &88.0  & 65.7& 88.6 &93.8 &  75.6& 94.1& 96.8& 87.3& 98.0 &99.2  \\ 
    VILLA$_L$~\cite{gan2020large} & \scalebox{0.98}{Same as Test}& -& - &- & -& - &- & 76.3 &94.2 &96.8 & 87.9 &97.5 &98.8  \\     
    Oscar$_L$~\cite{li2020oscar} & \scalebox{0.98}{Same as Test}& \underline{57.5} &\underline{82.8}& \textbf{89.8} & \underline{73.5}& \underline{92.2} &\underline{96.0} & -& -& -  & - &- &-   \\ 
    ERNIE-ViL$_L$~\cite{yu2021ernie} & \scalebox{0.98}{Same as Test}& - &-& - & - &- &- & 76.7 &93.6 &96.4& 88.7 &98.0& 99.2 \\     \midrule   
     CLIP$_B$~\cite{clip} &\xmark & 33.1 &58.4 &69.0 &52.5 & 76.7 & 84.7& 62.1& 85.7 & 91.9 & 82.2 & 96.6 & 99.0 \\ 
     FLYP$_B$~\cite{flyp} & COCO& 51.7 & 77.6 & 86.0& 69.7& 88.7& 93.9 & 76.3& 94.2& 96.8&89.0& 98.2& {99.5}   \\
     WiSE-FLYP$_B$~\cite{flyp}& COCO & 52.3& 77.7& 85.8& 70.3&89.3 &94.0 &77.3 &94.6 &{97.2} &91.0 &98.6 &99.3 \\
     \ccol  Ours$_B$ & \ccol COCO & \ccol {53.5} &\ccol {79.0}&\ccol {87.0} &\ccol {71.6} & \ccol {90.2}& \ccol {94.4} & \ccol \underline{78.4} & \ccol \underline{95.0}& \ccol \underline{97.5} &\ccol \underline{91.9}& \ccol \underline{98.7}&\ccol \textbf{99.6} \\
     \ccol  Ours$_L$ & \ccol COCO & \ccol \textbf{58.1} &\ccol \textbf{58.1}&\ccol \underline{89.0} &\ccol  \textbf{75.8} & \ccol \textbf{92.9}& \ccol \textbf{96.2} & \ccol \textbf{83.4} & \ccol \textbf{96.9}& \ccol \textbf{98.6} &\ccol \textbf{95.9}& \ccol \textbf{99.4}&\ccol \textbf{99.6} \\ 
\bottomrule 
\end{tabularx}
}
\label{tab:Retrieval}
\end{table*}

\subsection{Cross-Modal Retrieval}\label{sec:cross_retrieval}
We evaluate our model on COCO~\cite{Mscoco} and Flickr30K~\cite{Flickr30k_b} for cross-modal retrieval, where the model is only fine-tuned on the COCO dataset.
Since most previous methods for robust fine-tuning are limited to the classification task only, we compare our method with FLYP~\cite{flyp} from our re-implementation.
We further compare ours with supervised specialists~\cite{li2020unicoder, chen2020uniter, gan2020large, li2020oscar, yu2021ernie}.
As shown in Table~\ref{tab:Retrieval}, our method outperforms FLYP and its weight-ensemble~(WiSE-FLYP) in terms of all evaluation metrics both on COCO and Flickr30K.
Moreover, our method using CLIP ViT-L/14 surpasses the supervised specialists that have a similar size and are trained on both datasets, respectively.
Note that although we do not utilize Flickr30K in training, it outperforms supervised methods.

\setlength{\floatsep}{10pt plus 0pt minus 5pt}
\setlength{\textfloatsep}{13pt plus 0pt minus 5pt}
\begin{table}[!t]
\caption{Comparison of mIoU results between the OVSeg fine-tuned with our method and existing open-vocabulary segmentation models. Note that OVSeg (Org.) is trained in two stages, starting with full CLIP model fine-tuning followed by mask prompt tuning, whereas OVSeg (Ours) involves single-stage adapter training.}
\setlength{\tabcolsep}{1pt}
\fontsize{6.6}{8.6}\selectfont
\centering
\scalebox{0.98}
{
\begin{tabularx}{\columnwidth}
{
  p{0.18\textwidth}
  >{\centering\arraybackslash}p{0.155\textwidth} 
  >{\centering\arraybackslash}X
  >{\centering\arraybackslash}X
  >{\centering\arraybackslash}X
  >{\centering\arraybackslash}X 
  >{\centering\arraybackslash}X
  }
     \toprule
    \textbf{Methods}& \textbf{Backbone}&
   {{\textbf{A-847}}} & {{\textbf{PC-459}}}& {{\textbf{A-150}}}& {{\textbf{A-59}}} &{{\textbf{PAS-20}}} \\  \midrule
    ZegFormer~\cite{ding2022decoupling} & R-50~\cite{resnet} &- &- &16.4 &- &80.7 \\
    OpenSeg~\cite{ghiasi2022scaling} & R-101~\cite{resnet}&4.0 &6.5 &15.3& 36.9& 60.0  \\\midrule
     LSeg+~\cite{ghiasi2022scaling} & Eff-B7~\cite{tan2019efficientnet}  &3.8 &7.8 &18.0 &46.5 & - \\
     OpenSeg~\cite{ghiasi2022scaling} & Eff-B7~\cite{tan2019efficientnet}  &6.3 &9.0 &21.1 &42.1 &- \\ \midrule
     OVSeg (Org.)~\cite{ovseg} &   Swin-B~\cite{liu2021swin} &  \underline{9.0} &  
    \underline{12.4}&   \textbf{29.6}&  \underline{55.7}&  \underline{94.5}   \\
     \ccol OVSeg (Ours) & \ccol  Swin-B~\cite{liu2021swin} &  \ccol \textbf{10.3} \textcolor{grn}{($\uparrow$ 1.3)}  & \ccol \textbf{12.8} \textcolor{grn}{($\uparrow$ 0.4)}& \ccol \underline{29.5} \textcolor{red}{($\downarrow$ 0.1)}& \ccol \textbf{58.4} \textcolor{grn}{($\uparrow$ 2.7)}& \ccol \textbf{96.4} \textcolor{grn}{($\uparrow$ 1.9)}\\
\bottomrule 
\end{tabularx}
}
\label{tab:ovseg}
\end{table}

\subsection{Open-Vocabulary Segmentation}\label{sec:ovseg}

Our method can enhance open-vocabulary segmentation performance when used for fine-tuning the CLIP model within the OVSeg framework~\cite{ovseg}. We use full-rank adapters in the CLIP image and text encoders of OVSeg, fine-tuning them while keeping the pre-trained encoders frozen. Following the OVSeg setup, we employ MaskFormer~\cite{maskformer} with Swin-B~\cite{liu2021swin} as a mask proposal network, trained on the COCO-Stuff dataset~\cite{caesar2018coco}. 
The masked image classification model using CLIP ViT-L/14 is trained on masked images from COCO Captions~\cite{chen2015microsoft} with our method and evaluated on five unseen datasets, as shown in Table~\ref{tab:ovseg}. Compared to the original OVSeg model, OVSeg model fine-tuned with our method shows significant performance improvements, with mIoU increases of 1.3\%, 0.4\%, 2.6\%, and 1.9\% on A-847, PC-459, PC-59, and PAS-20, respectively. These results confirm that our method enhances generalization for unseen classes, showing it to be a promising approach for the open-vocabulary segmentation task.




\section{Conclusion}
\label{sec:conclusion}
We have introduced a novel approach for fine-tuning image-text models, emphasizing parameter efficiency and robustness to out-of-distribution data. By incorporating R-Adapter with self-ensembling techniques and MPM-NCE loss function, our method surpasses existing methods in robustness and efficiency. Moreover, its adaptability is confirmed by its successful application to diverse tasks. 
We believe that our method will greatly facilitate making the fine-tuning of zero-shot models much more broadly and easily accessible.

\section*{Acknowledgements}{
This work was supported by NRF grants (NRF-2021R1A2C3012728--30\%, NRF-2018R1A5A1060031--30\%, RS-2024-00341514--25\%) and IITP grants (RS-2019-II191906--10\%, Artificial Intelligence Graduate School Program - POSTECH, RS-2019-II190079--5\%, Artificial Intelligence Graduate School Program - Korea University) funded by Ministry of Science and ICT, Korea.
}

%
%
\bibliographystyle{splncs04}
\bibliography{cvlab_kwak}

\newpage
\setcounter{section}{0}
\renewcommand\thesection{\Alph{section}}
\renewcommand*{\theHsection}{\thesection}

\section{Implementation Details}

\noindent \textbf{Training.} 
The details of training configurations for full-/16-shot image classification, cross-modal retrieval, and open-vocabulary segmentation are presented in Table~\ref{tab:config}.
Moreover, Table~\ref{tab:config_generalization} presents the detailed training configuration for image classification in base-to-novel generalization~(\ref{suppsec:gen}).

\noindent \textbf{Image Augmentation.} Following CLIP~\cite{clip} and the previous work~\cite{wiseft, flyp}, training images are randomly cropped to match the default pixel resolution of the model (e.g., 224$\times$224 or 336$\times$336), without employing additional data augmentation techniques. For testing, images are simply resized to default image sizes.

\noindent \textbf{Text Templates.} 
For image classification tasks, regardless of the dataset, we utilize the 80 text templates related to ImageNet as proposed in CLIP~\cite{clip}. In the full-shot learning setting, during training, we randomly sample one of the text templates to construct the text following FLYP~\cite{flyp}. For few-shot learning, we primarily use the single text template, ``\texttt{a photo of a \{class\}}'', following CoOp~\cite{coop} and CoCoOp~\cite{cocoop}.
During the evaluation, we construct the classifier weights by employing an ensemble of prompts generated from the 80 text templates to construct the classifier weights, following CLIP, WiSE-FT~\cite{wiseft}, and FLYP.


\begin{table}[!b]
    \caption{Training configurations of various tasks.}
    \fontsize{8.5}{10.5}\selectfont
    \setlength{\tabcolsep}{4pt}
    \centering
    \scalebox{0.84}{
    \begin{tabular}{l|c c c c}
    \toprule
    \multirow{2}{*}{Configuration}& Classification & Classification & Cross-Modal   & Open-Vocabulary \\
    & (full-shot) & (16-shot) & Retrieval & Segmentation \\  \midrule
    Source dataset & ImageNet-1K~\cite{Imagenet} & ImageNet-1K~\cite{Imagenet}  & COCO~\cite{Mscoco}& COCO Captions~\cite{chen2015microsoft} \\\midrule
    \multirow{2}{*}{Image encoder} & 3 CLIP ViTs & \multirow{2}{*}{CLIP ViT-B/16} & 2 CLIP ViTs & \multirow{2}{*}{CLIP ViT-L/14} \\
    & (B/32, B/16, L/14@336px) & & (B/16, L/14) &\\ \midrule
    Batch size & 512 & 256 & 512 & 256 \\ 
    Total epochs & 10 & 50 & 10 & 5 \\
    Optimizer & \multicolumn{4}{c}{AdamW~\cite{adamw}} \\
    Scheduler & \multicolumn{4}{c}{Cosine-annealing schedule~\cite{loshchilov2016sgdr}} \\ 
    Warm-up step & \multicolumn{4}{c}{500} \\
    Initial learning rate & \multicolumn{4}{c}{$\expnum{5}{4}$} \\
    Drop Probability $p$ &\multicolumn{4}{c}{0.2}\\
    Momentum $m$ & \multicolumn{4}{c}{0.999}   \\
    Temperature $\tau$ & \multicolumn{4}{c}{0.01}  \\
    Margin $\delta$ & \multicolumn{4}{c}{0.05} \\    
    Label Smoothing Noise $\epsilon$ & 0.05 & 0 & 0 & 0  \\  
    Re-scaling coefficient $\alpha$ & 0.5 & 0.5 & 0.8 & 0.4 \\ 
    \bottomrule
    \end{tabular}
    }
	\label{tab:config}   
\end{table}

    

\noindent \textbf{Open-Vocabulary Segmentation.}
By following~\cite{ovseg}, the original OVSeg model consists of two components. 
One is a mask proposal network \ie, MaskFormer~\cite{maskformer}, and the other is the CLIP image and text encoders~\cite{clip}.
Specifically, the mask proposal network with Swin-B~\cite{liu2021swin} as a backbone pre-trained on the COCO-Stuff dataset~\cite{caesar2018coco} produces several segmentation masks given an image input.
Meanwhile, the CLIP image encoder is trained on the image masks from COCO Captions~\cite{chen2015microsoft} in two stages, starting with full fine-tuning followed by mask prompt tuning, while freezing the CLIP text encoder.
OVSeg employs a dataset composed of mask proposals from MaskFormer and their predictions, which is used for training. 
In contrast, we adopt a training strategy for OVSeg that is significantly different from the original training strategy.
OVSeg with our method involves a \textbf{single-stage training process} focused solely on our adapters in the CLIP model. We utilize the ground truth masks and categories from COCO Caption for training. This approach was initially suggested to result in performance degradation in the original OVSeg paper (as mentioned in Table 2 of OVSeg paper~\cite{ovseg}). In our implementation, our method overcomes the issues they identified and achieves even higher performance.
Interestingly, we found that the performance degraded when mask prompt tuning was used in conjunction with our method.
For testing, the final class predictions are computed by an ensemble of the prediction of MaskFormer model and the prediction of CLIP model following the same setting of OVSeg.
Specifically, when the prediction weight of CLIP is denoted as $x$ and the prediction weight of MaskFormer as $y$, the ensemble is expressed as $y^{(1-\lambda)}*x^\lambda$. For the ensemble value $\lambda$, we used 0.8 in A-847, 0.75 in PC-459, 0.8 in A-150, 0.5 in PC-59, and 0.25 in PAS-20.

\begin{table}[!t]
\caption{Training configurations of image classification in base-to-novel generalization setting.}
    \fontsize{8.5}{10.5}\selectfont
    \setlength{\tabcolsep}{6pt}
    \centering
    {\begin{tabular}{l|c}
    \toprule
    \multirow{2}{*}{Configuration}& Classification \\ & (Base-to-Novel) \\  \midrule
    
    \multirow{1}{*}{Image encoder} & {CLIP ViT-B/16} \\\midrule
    
    Batch size & 32  \\ 
    Total epochs & 100  \\
    Optimizer & \multicolumn{1}{c}{AdamW~\cite{adamw}} \\
    Scheduler & \multicolumn{1}{c}{Cosine~\cite{loshchilov2016sgdr}} \\ 
    Warm-up step & \multicolumn{1}{c}{500} \\
    Initial learning rate & \multicolumn{1}{c}{$\expnum{5}{4}$} \\
    Drop Probability $p$ &\multicolumn{1}{c}{0.2}\\
    Momentum $m$ & \multicolumn{1}{c}{0.9}   \\
    Temperature $\tau$ & \multicolumn{1}{c}{0.01}  \\
    Margin $\delta$ & \multicolumn{1}{c}{0.05} \\    
    Label Smoothing Noise $\epsilon$ & 0   \\  
    Re-scaling coefficient $\alpha$ & 0.5  \\ 
    \bottomrule
    \end{tabular}}
	\label{tab:config_generalization}
\end{table}

\section{Datasets Details}
\noindent{\textbf{Image Classification.}} We use ImageNet (IN)~\cite{Imagenet} as the ID dataset for fine-tuning; we evaluate the robustness of the models on five standard OOD datasets that represent five different types of OOD scenarios:
ImageNetV2 (IN-V2)~\cite{recht2019imagenet} is a new test set for ImageNet with distribution shift.
ImageNet-R (IN-R)~\cite{hendrycks2021many} consists of various artistic renditions (\eg, painting, cartoons) of 200 ImageNet classes
ImageNet-Sketch (IN-Sketch)~\cite{wang2019learning} contains sketch images of 1000 ImageNet classes. 
ObjectNet~\cite{barbu2019objectnet} is a test set that contains images with 313 object classes collected from new viewpoints on new backgrounds, where 113 classes overlap with ImageNet. 
ImageNet-A (IN-A)~\cite{hendrycks2021natural} consists of natural images that are misclassified by a pre-trained ResNet-50~\cite{resnet} for 200 ImageNet classes.

\noindent{\textbf{Cross-Modal Retrieval.}}
We utilize two standard benchmarks for image-text cross-modal retrieval, COCO~\cite{Mscoco} as ID and Flickr30K~\cite{Flickr30k_b} as OOD. For these two datasets, each image is associated with the corresponding five captions.
Specifically, COCO is exploited as the ID dataset, and Flickr30K is utilized as the OOD dataset which has distribution shifts in both image and text modalities. 
In COCO, there are 123,287 images, and we follow the data split of~\cite{karpathy2015deep} with 113,287 images for training, and 5,000 images for testing.
Flickr30K contains 29,000 images for training and 1,000 images for testing.

\noindent{\textbf{Open-Vocabulary Segmentation.}}
By following~\cite{ovseg}, we train the models on COCO Captions~\cite{chen2015microsoft} and evaluate them on ADE20K~\cite{zhou2019semantic}, Pascal Context~\cite{mottaghi2014role}, and Pascal VOC~\cite{Pascalvoc} with 20 categories (PAS-20).
Specifically, we exploit ADE20K in two versions, one with 150 frequently used categories (A-150) and the other with diverse 847 categories (A-847).
Moreover, we also utilize Pascal Context in two versions, one with 59 frequently used categories (PC-59) and the other with the whole 459 categories (PC-459).
Following our baseline method~\cite{ovseg}, we train a CLIP model on the COCO Captions dataset \cite{chen2015microsoft} and test them on several benchmarks as OOD: ADE20K \cite{zhou2019semantic}, Pascal Context \cite{mottaghi2014role}, and Pascal VOC \cite{Pascalvoc}. 

\noindent{\textbf{Image Classification in Base-to-Novel Generalization.}} In our study of base-to-novel generalization for image classification, we employed 11 image recognition datasets as used in CoOp~\cite{coop}, encompassing a wide range of recognition tasks. The benchmark includes: ImageNet~\cite{Imagenet} and Caltech101~\cite{caltech101} for generic object classification; OxfordPets~\cite{oxfordpets}, StanfordCars~\cite{krause20133d}, Flowers102~\cite{flowers102}, Food101~\cite{food101}, and FGVCAircraft~\cite{fgvcaircraft} for fine-grained classification; SUN397~\cite{sun397} for scene recognition; UCF101~\cite{ucf101} for action recognition; DTD~\cite{dtd} for texture classification; and EuroSAT~\cite{eurosat} for satellite imagery recognition.

\section{Additional Experiments}
\subsection{Generalization From Base to Novel Classes}\label{suppsec:gen}

We conduct experiments to further emphasize generalizability by utilizing 11 datasets to measure the generalization performance in a base-to-novel setting following CoCoOp~\cite{cocoop}.
On each of the 11 datasets, we divide the classes into two equal groups: base classes and novel classes. All models are trained using only the base classes, with 16 samples per class, while evaluation is conducted on both base and novel classes separately to test generalizability.

As a default setting for training on few-shot datasets, we construct a text description of the target class employing a single text-template. We experiment with two settings varying bottleneck dimensions of R-Adapter: 4-rank and full-rank, with fewer and more parameters, respectively.
We further explore the model when employing a full-rank structure and sampling templates from a predefined set of multiple text templates.
The results are shown in Table~\ref{tab:basetonew}.

\begin{table}[!h]
\caption{Comparison with fine-tuned methods from CLIP in base-to-novel generalization. All methods are trained from the base classes (16 shots). HM denotes Harmonic mean~\cite{xian2017zero} which emphasizes the generalization trade-off. Superscripts denote the rank of adapter modules. ``MT'' represents that text description of the target class is constructed by sampling from a set of multiple predefined templates as used in FLYP~\cite{flyp}.
}
    \centering
    \fontsize{7}{9}\selectfont
    \setlength{\tabcolsep}{2pt}

    \begin{subtable}[t]{0.325\linewidth}
        \centering
        \resizebox{1.0\linewidth}{!}{
        \begin{tabular}{l|cc|c}
        \toprule
         & Base & Novel & HM \\
        \midrule
        CLIP~\cite{clip} &69.34 & 74.22& 71.70\\
        CoOp~\cite{coop} & 82.63& 67.99& 74.60\\
        CoCoOp~\cite{cocoop} & 80.47& 71.69&75.73 \\
        KgCoOp~\cite{coop} & 80.73& 73.60& 77.00\\
        MaPLe~\cite{maple}  & 82.28& 75.14& 78.55 \\ \midrule
        Ours$^4$ & 80.06& \textbf{76.27}&78.11 \\
        Ours$^{8}$ & 81.74 & 76.45& 79.01  \\
        Ours$^{16}$ & 82.34& 76.25 & 79.18 \\
        Ours$^{32}$ & 83.00 & 76.16 & 79.43\\
        Ours$^\textrm{Full}$ &83.21 & {75.82}& {79.34}\\
        Ours$^\textrm{Full}$ (MT) &\textbf{83.64} & 76.08& \textbf{79.68}\\
        \bottomrule
        \end{tabular}
        }
        \caption{{Average over 11 datasets}}
        \label{tab:avg}
    \end{subtable}
    \hfill
    \begin{subtable}[t]{0.325\linewidth}
        \centering
        \resizebox{1.0\linewidth}{!}{
        \begin{tabular}{l|cc|c}
        \toprule
         & Base & Novel & HM \\
        \midrule
        CLIP~\cite{clip} & 72.43& 68.14& 70.22\\
        CoOp~\cite{coop} &76.46 &66.31 & 71.02\\
        CoCoOp~\cite{cocoop} &75.98 & 70.43& 73.10\\
        KgCoOp~\cite{coop} & 75.73& 69.96& 72.78\\
        MaPLe~\cite{maple} & 76.66 &70.54 & 73.47 \\ \midrule
        Ours$^4$ & 76.38& 71.38&73.87 \\
        Ours$^{8}$ & 76.39& 71.81& 74.03  \\
        Ours$^{16}$ & 76.38 & 71.58& 73.90 \\
        Ours$^{32}$ & 76.76& 71.64& 74.11\\
        Ours$^\textrm{Full}$ & {77.57}&{71.58} & {74.46}\\
        Ours$^\textrm{Full}$ (MT) &\textbf{77.74} & \textbf{71.70}& \textbf{74.60}\\
        \bottomrule
        \end{tabular}
        }
        \caption{ImageNet}
        \label{tab:ImageNet}
    \end{subtable}
    \hfill
    \begin{subtable}[t]{0.325\linewidth}
        \centering
        \resizebox{1.0\linewidth}{!}{
        \begin{tabular}{l|cc|c}
        \toprule
         & Base & Novel & HM \\
        \midrule
        CLIP~\cite{clip} &96.84& 94.00& 94.50\\
        CoOp~\cite{coop} &98.11 &93.52 & 95.76\\
        CoCoOp~\cite{cocoop} &97.96 &93.81 &95.84 \\
        KgCoOp~\cite{coop} & 97.72& 94.39& 96.03\\ 
        MaPLe~\cite{maple} & 97.74 & 94.36& 96.02 \\ \midrule
        Ours$^4$ &97.74 & {95.85} & 96.79\\
        Ours$^{8}$ &98.44 &96.02 &97.22  \\
        Ours$^{16}$ & 98.21& 96.19& 97.19\\
        Ours$^{32}$ & 98.67& 95.77& 97.20 \\
        Ours$^\textrm{Full}$ &\textbf{98.83} &{95.67} &{97.23} \\
        Ours$^\textrm{Full}$ (MT) & 98.21& \textbf{96.36}& \textbf{97.28}\\
        \bottomrule
        \end{tabular}
        }
        \caption{Caltech101}
        \label{tab:Caltech101}
    \end{subtable}

    \begin{subtable}[b]{0.325\linewidth}
        \centering
        \resizebox{1.0\linewidth}{!}{
        \begin{tabular}{l|cc|c}
        \toprule
         & Base & Novel & HM \\
        \midrule
        CLIP~\cite{clip} & 91.17& 97.26& 94.12\\
        CoOp~\cite{coop} &94.24 &96.66 &95.43 \\
        CoCoOp~\cite{cocoop} &95.20 &97.69 &96.43 \\
        KgCoOp~\cite{coop} & 94.65& \textbf{97.76}& 96.18\\ 
        MaPLe~\cite{maple} & 95.43 &\textbf{97.76} &{96.58} \\ \midrule
        Ours$^4$ &93.73& 97.71& 95.68\\
        Ours$^{8}$ &95.75 &96.92 &96.33 \\
        Ours$^{16}$ &95.96 & 98.04&96.99 \\
        Ours$^{32}$ & 95.91& 97.54& 96.72\\
        Ours$^\textrm{Full}$ & \textbf{95.80}& 97.37&{96.58} \\
        Ours$^\textrm{Full}$ (MT) & 96.65& 97.48&\textbf{97.07} \\
        \bottomrule
        \end{tabular}
        }
        \caption{OxfordPets}
        \label{tab:OxfordPets}
    \end{subtable}
    \hfill
    \begin{subtable}[b]{0.325\linewidth}
        \centering
        \resizebox{1.0\linewidth}{!}{
        \begin{tabular}{l|cc|c}
        \toprule
         & Base & Novel & HM \\
        \midrule
        CLIP~\cite{clip} & 63.37&74.89 &68.65 \\
        CoOp~\cite{coop} & 76.20&60.40 &72.49 \\
        CoCoOp~\cite{cocoop} & 70.49& 73.59& 72.01\\
        KgCoOp~\cite{coop} & 71.76& 75.04& 73.36\\ 
        MaPLe~\cite{maple} & 72.94 &74.00 &73.47 \\ \midrule
        Ours$^4$ & 79.11& 74.85&76.92 \\
        Ours$^{8}$ &78.74& 75.58& 77.12 \\
        Ours$^{16}$ & 77.87 & 75.05& 76.44\\
        Ours$^{32}$ & 78.91& 75.88 & 77.36 \\
        Ours$^\textrm{Full}$ & {81.24}& \textbf{75.98}&\textbf{78.52} \\
        Ours$^\textrm{Full}$ (MT) & \textbf{81.88}& 74.15&77.82\\
        \bottomrule
        \end{tabular}
        }
        \caption{StanfordCars}
        \label{tab:StanfordCars}
    \end{subtable}
    \hfill
    \begin{subtable}[b]{0.325\linewidth}
        \centering
        \resizebox{1.0\linewidth}{!}{
        \begin{tabular}{l|cc|c}
        \toprule
         & Base & Novel & HM \\
        \midrule
        CLIP~\cite{clip} & 72.08& 77.80& 74.83\\
        CoOp~\cite{coop} &97.63 &69.55 &81.23 \\
        CoCoOp~\cite{cocoop} & 94.87& 71.75&81.71 \\
        KgCoOp~\cite{coop} & 95.00& \textbf{74.73}& \textbf{83.65}\\ 
        MaPLe~\cite{maple} & \textbf{95.92} &72.46& 82.56 \\ \midrule
        Ours$^4$ & 87.34& 74.03& 80.14\\
        Ours$^{8}$ & 91.42& 72.63& 80.95 \\
        Ours$^{16}$ & 91.33& 73.79&81.63 \\
        Ours$^{32}$ & 92.86& 74.34&82.57 \\
        Ours$^\textrm{Full}$ &90.09 & 73.25& 81.16\\
        Ours$^\textrm{Full}$ (MT) & 95.07 & 73.56& 82.94\\
        \bottomrule
        \end{tabular}
        }
        \caption{Flowers102}
        \label{tab:Flowers102}
    \end{subtable}

    \begin{subtable}[b]{0.325\linewidth}
        \centering
        \resizebox{1.0\linewidth}{!}{
        \begin{tabular}{l|cc|c}
        \toprule
         & Base & Novel & HM \\
        \midrule
        CLIP~\cite{clip} &90.10 &91.22 &90.66 \\
        CoOp~\cite{coop} &89.44 &87.50 &88.46 \\
        CoCoOp~\cite{cocoop} & 90.70&91.29 &90.99 \\
        KgCoOp~\cite{coop} & 90.50&91.70 &91.90 \\ 
        MaPLe~\cite{maple} & \textbf{90.71}&\textbf{92.05}& \textbf{91.38} \\ \midrule
        Ours$^4$ & 90.28& 90.79& 90.54\\
        Ours$^{8}$ &90.50 & 91.31& 90.9 \\
        Ours$^{16}$ & 90.58& 91.33 &90.95 \\
        Ours$^{32}$ & 90.55& 91.42& 90.98 \\
        Ours$^\textrm{Full}$ & 90.29& 90.05&90.17 \\
        Ours$^\textrm{Full}$ (MT) & 90.46& 91.33& 90.89\\
        \bottomrule
        \end{tabular}
        }
        \caption{Food101}
        \label{tab:Food101}
    \end{subtable}
    \hfill
    \begin{subtable}[b]{0.325\linewidth}
        \centering
        \resizebox{1.0\linewidth}{!}{
        \begin{tabular}{l|cc|c}
        \toprule
         & Base & Novel & HM \\
        \midrule
        CLIP~\cite{clip} & 27.19&36.29 &31.09 \\
        CoOp~\cite{coop} & 39.24&30.49 &34.30 \\
        CoCoOp~\cite{cocoop} & 33.41&23.71 &27.74 \\
        KgCoOp~\cite{coop} & 36.21& 35.55& 34.83\\ 
        MaPLe~\cite{maple} & 37.44& 35.61& 36.50\\ \midrule
        Ours$^4$ & 36.01&\textbf{37.07} &36.54 \\
        Ours$^{8}$ & 35.71& 37.55&36.61  \\
        Ours$^{16}$ & 39.20& 36.71& 37.91\\
        Ours$^{32}$ & 39.56& 35.33& 37.33\\
        Ours$^\textrm{Full}$ & \textbf{41.48} & 36.17& \textbf{38.64}\\
        Ours$^\textrm{Full}$ (MT) & 40.04 & 35.73& 37.77\\
        \bottomrule
        \end{tabular}
        }
        \caption{FGVCAircraft}
        \label{tab:FGVCAircraft}
    \end{subtable}
    \hfill
    \begin{subtable}[b]{0.325\linewidth}
        \centering
        \resizebox{1.0\linewidth}{!}{
        \begin{tabular}{l|cc|c}
        \toprule
         & Base & Novel & HM \\
        \midrule
        CLIP~\cite{clip} & 69.36&75.35 &72.23 \\
        CoOp~\cite{coop} & 80.85& 68.34&74.07 \\
        CoCoOp~\cite{cocoop} & 79.74&76.86 &78.27 \\
        KgCoOp~\cite{coop} & 80.29&76.53 &78.36 \\ 
        MaPLe~\cite{maple} & 80.82& \textbf{78.70}& {79.75}\\ \midrule
        Ours$^4$ & 80.42 & 78.43 & 79.42 \\
        Ours$^{8}$ & 81.41& 78.63& 79.99 \\
        Ours$^{16}$ & 81.76& 77.98& 79.82\\
        Ours$^{32}$ & 82.08& 78.24& 80.12 \\
        Ours$^\textrm{Full}$ & {81.38}& 78.06& 79.68\\
        Ours$^\textrm{Full}$ (MT) & \textbf{82.70}& 78.36& \textbf{80.48}\\
        \bottomrule
        \end{tabular}
        }
        \caption{SUN397}
        \label{tab:SUN397}
    \end{subtable}

    \begin{subtable}[b]{0.325\linewidth}
        \centering
        \resizebox{1.0\linewidth}{!}{
        \begin{tabular}{l|cc|c}
        \toprule
         & Base & Novel & HM \\
        \midrule
        CLIP~\cite{clip} & 53.24& 59.90& 56.37\\
        CoOp~\cite{coop} & 80.17& 47.54& 59.68\\
        CoCoOp~\cite{cocoop} & 77.01& 56.00& 64.85\\
        KgCoOp~\cite{coop} & 77.55& 54.99& 64.35\\ 
        MaPLe~\cite{maple} & 80.36 &59.18 &68.16\\ \midrule
        Ours$^4$ & 73.15& \textbf{66.43}& 69.62\\
        Ours$^{8}$ & 77.43& 66.91& 71.79  \\
        Ours$^{16}$ &79.05& 63.89&70.67 \\
        Ours$^{32}$ & 79.86& 64.37&  71.28\\
        Ours$^\textrm{Full}$ & \textbf{83.45}& 64.13& {72.53}\\
        Ours$^\textrm{Full}$ (MT) & 83.33& 64.62&\textbf{72.79} \\
        \bottomrule
        \end{tabular}
        }
        \caption{DTD}
        \label{tab:DTD}
    \end{subtable}
    \hfill
    \begin{subtable}[b]{0.325\linewidth}
        \centering
        \resizebox{1.0\linewidth}{!}{
        \begin{tabular}{l|cc|c}
        \toprule
         & Base & Novel & HM \\
        \midrule
        CLIP~\cite{clip} & 56.48& 64.05& 60.03\\
        CoOp~\cite{coop} & 91.54& 54.44& 68.27\\
        CoCoOp~\cite{cocoop} & 87.49& 60.04& 71.21\\
        KgCoOp~\cite{coop} & 85.64& 64.34& 73.48\\ 
        MaPLe~\cite{maple} & \textbf{94.07} &73.23& \textbf{82.35}\\ \midrule
        Ours$^4$ & 84.88 & \textbf{75.85} & 80.11\\
        Ours$^{8}$ & 90.33&76.03 &82.56  \\
        Ours$^{16}$ & 90.74& 75.54& 82.44\\
        Ours$^{32}$ & 92.74& 74.79& 82.81\\
        Ours$^\textrm{Full}$ & 90.14 & 74.26 & 81.43\\
        Ours$^\textrm{Full}$ (MT) & 88.74& 74.92& 81.25\\
        \bottomrule
        \end{tabular}
        }
        \caption{EuroSAT}
        \label{tab:EuroSAT}
    \end{subtable}
    \hfill
    \begin{subtable}[b]{0.325\linewidth}
        \centering
        \resizebox{1.0\linewidth}{!}{
        \begin{tabular}{l|cc|c}
        \toprule
         & Base & Novel & HM \\
        \midrule
        CLIP~\cite{clip} &70.53 &77.50 & 73.85\\
        CoOp~\cite{coop} & 85.14& 64.47&73.37 \\
        CoCoOp~\cite{cocoop} & 82.33& 73.45& 77.64\\
        KgCoOp~\cite{coop} & 82.89& 76.67& 79.65\\ 
        MaPLe~\cite{maple} & 83.00& \textbf{78.66}& 80.77 \\ \midrule
        Ours$^4$ & 81.64& 76.58&79.03 \\
        Ours$^{8}$ & 83.04& 77.61& 80.23 \\
        Ours$^{16}$ & 84.64& 78.69& 81.56\\
        Ours$^{32}$ & 85.06& 78.47& 81.63 \\
        Ours$^\textrm{Full}$ & {85.06}&77.45 &{81.07} \\
        Ours$^\textrm{Full}$ (MT) & \textbf{85.21} &  78.64& \textbf{81.79}\\
        \bottomrule
        \end{tabular}
        }
        \caption{UCF101}
        \label{tab:UCF101}
    \end{subtable}

	\label{tab:basetonew}
\end{table}
\clearpage

\noindent \textbf{Advantages in Performance.}
 Our model with full-rank significantly outperformed the existing state of the art on most datasets by a large margin. Our method shows an average improvement of more than 1\%p in base classes and over 0.7\%p in novel classes compared to existing methods. Additionally, our model with 4-rank, which has a similar number of parameters as existing methods, performed better on new classes compared to our full-rank one, clearly achieving state-of-the-art performance. Overall, in terms of harmonic mean, our method achieves higher performance than existing methods, except for MaPLe~\cite{maple}.
Moreover, we found that using a set of multiple text-templates for sampling and training, instead of a single text-template, resulted in even greater performance gains. Consequently, this approach yields a 1.13\%p improvement in the harmonic mean over the MaPLe, demonstrating the effectiveness of diversifying textual input during training.

\noindent \textbf{Advantages in Efficiency.}
Our method easily adjusts to the required number of parameters by controlling the bottleneck dimension, without any added latency during inference. However, all existing baseline methods increase inference latency with added parameters since they involve adding input sequences. Especially, MaPLe, which achieved state-of-the-art performance, adds prompts to both text and visual encoders, significantly increasing its inference latency. Considering these factors, our method is highlighted for maintaining the same amount of computation as the original pre-trained model while achieving state-of-the-art performance.

\setlength{\floatsep}{10pt plus 0pt minus 5pt}
\setlength{\textfloatsep}{15pt plus 0pt minus 5pt}
\begin{table}[!t]
\caption{
Harmonic mean accuracy on base and novel classes. All methods are fine-tuned with 16 shots per base class.
}
\label{tab:base_to_novel_ablation}
\centering
\setlength{\tabcolsep}{1pt}
\resizebox{\linewidth}{!}{
\fontsize{7}{9}\selectfont
\begin{tabularx}{\linewidth}
{
  p{0.02\textwidth} 
  > {\centering\arraybackslash}p{0.09\textwidth} |
  >{\centering\arraybackslash}X 
  >{\centering\arraybackslash}X
  >{\centering\arraybackslash}X
  >{\centering\arraybackslash}X
  >{\centering\arraybackslash}X
  >{\centering\arraybackslash}X
  >{\centering\arraybackslash}X
  >{\centering\arraybackslash}X
  >{\centering\arraybackslash}X
  >{\centering\arraybackslash}X
  >{\centering\arraybackslash}X
  >{\centering\arraybackslash}X
  }
 \toprule
\multicolumn{1}{l}{Methods}&
{\#Param} & Avg & {IN} &  {Cal} & {Pets}& {Cars} & {Flo} & {Food} & {Air} & {SUN} & {DTD} & {Euro} & {UCF}\\ \midrule
MaPLE & {3.55 M}& 78.6 & 73.5 & 96.0 & 96.6 & 73.5 & 82.6 & 91.4 & 36.5 & 79.8 & 68.2 & 82.4 & 80.8  \\ \midrule 
Ours$^4$ & 0.25 M& 78.1& 73.9 & 96.8 & 95.7 & 76.9 & 80.1 & 90.5 & 36.5 & 79.4 & 69.6 & 80.1 & 79.0  \\
Ours$^{8}$ & 0.49 M& 79.0& 74.0&\textbf{97.2} &96.3 &77.1 &81.0 &90.9 & 36.6& 80.0& \textbf{71.8} &82.6& 80.2    \\
Ours$^{16}$ & 0.98 M&79.2 & 73.9& \textbf{97.2}& \textbf{97.0}&76.4 &81.6 &81.0 &\textbf{37.9} &79.8 &70.7 & 82.4 & 81.5    \\
Ours$^{32}$ & 1.97 M& \textbf{79.4} & \textbf{74.1}& \textbf{97.2}& 96.7&\textbf{77.4} &\textbf{82.6} &\textbf{91.0} &37.3 &\textbf{80.1} &71.3 & \textbf{82.8} & \textbf{81.6}    \\
\bottomrule
\end{tabularx}
}
\end{table}

\subsection{Detailed Comparison to Parameter-Efficient Fine-Tuning}
In this analysis, we conduct a detailed comparison among parameter-efficient fine-tuning (PEFT) methods, specifically focusing on LoRA, AdaptFormer, and RepAdapter. It's important to recall that R-Adapter utilizes a bottleneck module consisting of two matrices when the adapter rank is smaller than the hidden dimension of the backbone encoder. Conversely, R-Adapter with a full-rank employs a singular matrix due to the omission of non-linear layers, leveraging a multiplicative bottleneck structure. In our experiment, regardless of methods, all adapter modules are uniformly attached to both image and text encoders, ensuring fairness. However, the attachment locations and attachment manner differ among the approaches, leading to variations in the number of parameters even at the same rank.

We note that as the rank increases across all methods, there is a corresponding increase in the number of parameters, which significantly enhances performance in ID data. However, all existing methods show a decrease in OOD generalization performance as rank increases. In contrast, our method demonstrates robustness in OOD even at lower ranks and, unlike other methods, shows an improvement in OOD performance as the rank increases, creating a substantial gap in OOD performance between our method and existing approaches. Consequently, when using a similar number of parameters, our method not only outperforms existing PEFT methods in terms of performance but also ensures robustness irrespective of rank.

\setlength{\floatsep}{10pt plus 0pt minus 5pt}
\setlength{\textfloatsep}{15pt plus 0pt minus 5pt}
\begin{table}[!t]
\caption{Top-1 accuracy of parameter-efficient fine-tuning methods on ImageNet (ID) and OOD datasets with ViT-B/32. Superscripts denote the rank of adapter or LoRA.
}
\fontsize{7}{9}\selectfont
\setlength{\tabcolsep}{2pt}
\centering
\scalebox{1.0}{
\begin{tabularx}{1.0\textwidth}
    {
      p{0.182\textwidth}
      >{\centering\arraybackslash}p{0.15\textwidth} 
      >{\centering\arraybackslash}p{0.065\textwidth} 
      >{\centering\arraybackslash}p{0.11\textwidth} |
      >{\centering\arraybackslash}p{0.065\textwidth}
      >{\centering\arraybackslash}p{0.065\textwidth}
      >{\centering\arraybackslash}p{0.1\textwidth}
      >{\centering\arraybackslash}p{0.1\textwidth}
      >{\centering\arraybackslash}p{0.065\textwidth}
      }
     \toprule
    {\multirow{1}{*}[-4.5mm]{\textbf{Methods}}}& \multirow{1}{*}[-0.5mm]{\textbf{Trainable}}&
    \multicolumn{1}{c}{\textbf{ID}} & \multicolumn{6}{c}{\textbf{Out-Of-Distribution~(OOD)}} \\ [-0.3ex]  \cmidrule(lr){3-3}  
    \cmidrule(lr){4-9} &\multirow{1}{*}[0.5mm]{\textbf{Params (M)}} & IN & {OOD avg.} & IN-V2 & IN-R&  {IN-Sketch} &{ObjectNet}& IN-A  \\  [-0.4ex] \midrule
     AdaptFormer$^\textrm{16}$~\cite{adaptformer} & 0.5 & \textbf{74.7}& 48.9&  64.3 & 63.8 &41.7 & 45.5 & 29.3 \\ 
     RepAdapter$^\textrm{16}$~\cite{repadapter} & 1.0 &  74.3 & 49.7 & 64.4& 65.1& 42.4& 46.0& 30.4  \\
       Ours$^\textrm{16}$ & 1.0 &    {74.5}&  \textbf{52.5} &  \textbf{65.1} &  \textbf{69.5} &  \textbf{45.8} &   \textbf{47.9} &  \textbf{34.0} \\ 
     \midrule
     AdaptFormer$^\textrm{128}$~\cite{adaptformer} &3.9 &  75.6 & 48.3 & 64.5 & 61.7 & 41.0 & 45.0 & 29.3 \\ 
     RepAdapter$^\textrm{128}$~\cite{repadapter} &7.8&   \textrm{76.3}& 48.9 & 65.2 & 62.7& 41.9& 45.7& 29.2 \\
       Ours$^\textrm{128}$ & 7.8 &  \textbf{76.7} &  \textbf{53.7}  & \textbf{66.9} &  \textbf{70.2} &  \textbf{47.1} &   \textbf{48.7} &  \textbf{35.5}  \\  \midrule
       LoRA$^\textrm{Full}$ &  163.6 & \textbf{78.0} &  {48.2} &  {66.2} & {60.0} &    {42.3}   &  {45.0} &  {27.4}   \\ 
     AdaptFormer$^\textrm{Full}$~\cite{adaptformer} & 20.5 & 77.2 & 48.5 & 66.4 & 60.6 & 42.2 & 45.2& 28.0\\
     RepAdapter$^\textrm{Full}$~\cite{repadapter} & 41.0 &  76.9 & 47.7 & 65.5& 60.1& 41.3& 44.2& 27.6 \\

       Ours$^\textrm{Full}$ &  20.5 & {77.7} &  \textbf{54.3}   &  \textbf{67.7} & \textbf{70.8} &    \textbf{47.8}   &  \textbf{49.7} &  \textbf{35.6}  \\ \bottomrule 
\end{tabularx}
}
\label{tab:petl_comparison}
\end{table}

\subsection{Additional Ablation Studies}

\noindent \textbf{Ablation Study on Re-scaling Coefficient.}
We investigate the impact of the re-scaling coefficient $\alpha$ in various tasks.
The effect varies with each task and dataset, and as the distribution shift between in-distribution (ID) and out-of-distribution (OOD) data increases, performance improvement is noted when the re-scaling parameter value is smaller. In ImageNet classification, as analyzed in WiSE-FT~\cite{wiseft}, fixing the scaling parameter to 0.5 yields sufficiently high performance for both ID and OOD data, and tuning it can achieve even higher performance. In Cross-modal Retrieval, although the distribution gap between COCO and Flickr30K is not very large, a continuous increase is observed as the scaling parameter increases. However, performance improvement is still noted compared to when scaling is not applied. 
In open vocabulary segmentation, we observe that the mIOU performance generally improves as the coefficient moderately increases, but it tends to decrease again when the coefficient becomes too large.

\begin{figure*}
     \centering
     \begin{subfigure}[b]{0.31\textwidth}
         \centering
         \includegraphics[width=\textwidth]{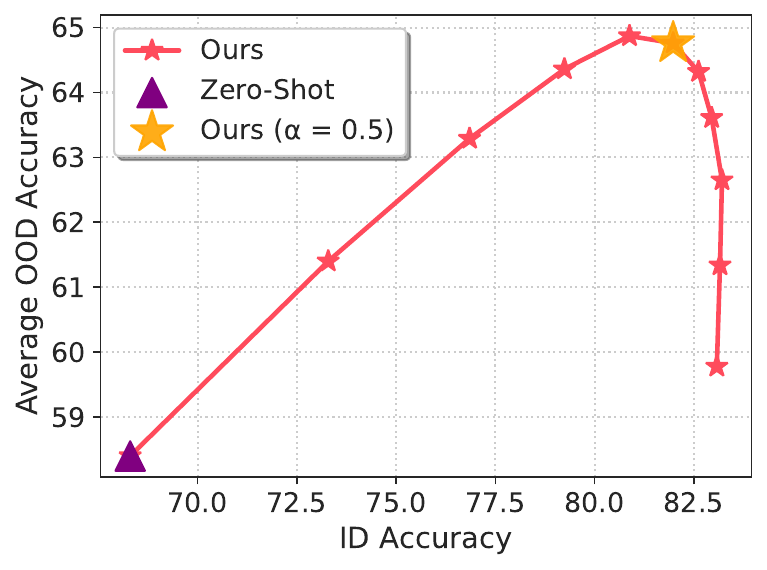}
         \caption{ImageNet Classification (ViT-B/16)}
     \end{subfigure}
     \hfill
     \begin{subfigure}[b]{0.33\textwidth}
         \centering
         \includegraphics[width=\textwidth]{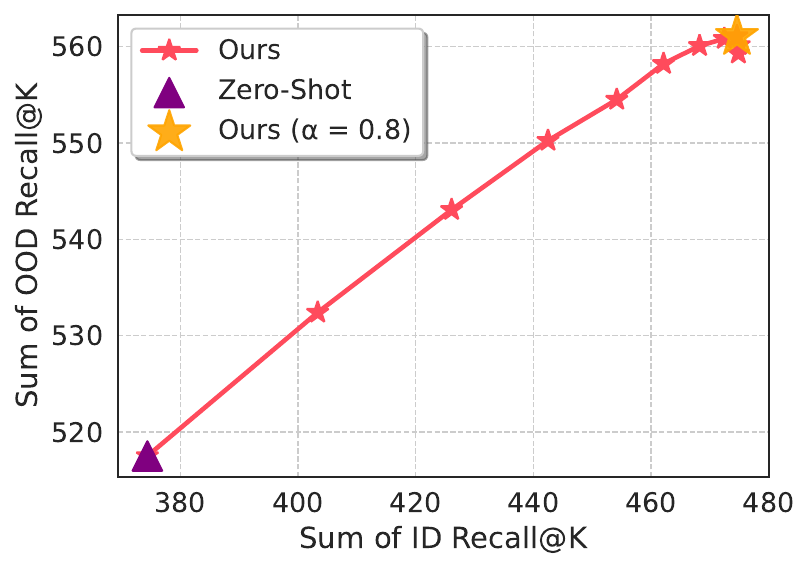}
         \caption{Cross-modal Retrieval (ViT-B/16)}
     \end{subfigure}
     \hfill
     \begin{subfigure}[b]{0.31\textwidth}
         \centering
         \includegraphics[width=\textwidth]{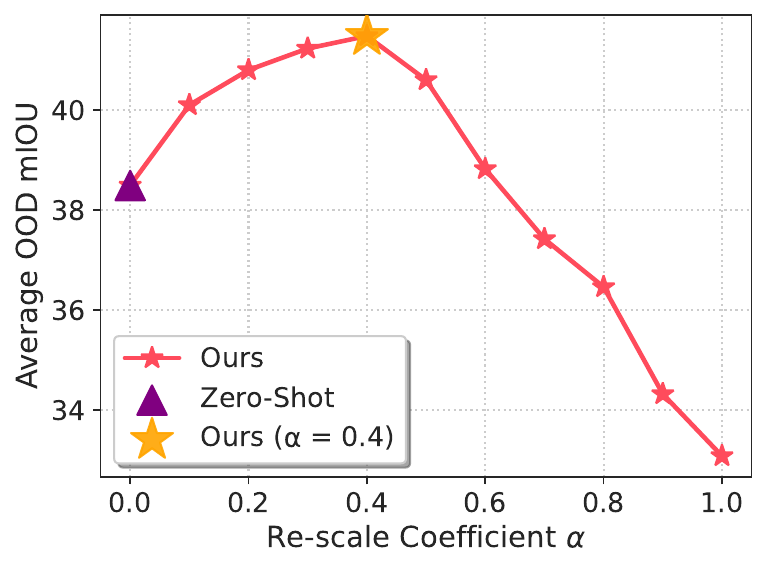}
         \caption{Open Vocabulary Segmentation (ViT-L/14)}
     \end{subfigure}
        \caption{Performance of our method varying re-scaling coefficient $\alpha$ in Eq.~9. The accuracy of each Cross-modal Retrieval is the sum of the performances in recall@K for Image retrieval (R@1, R@5, R@10) and the performances in recall@K for text retrieval (R@1, R@5, R@10).
        The accuracy of open vocabulary segmentation is the average of mIOU of 5 standard datasets.}
        \label{fig:varing_alpahs}
\end{figure*}

\begin{table}[h!]
\caption{Ablation study on label smoothing coefficient $\epsilon$ in Eq.~10.}
    \centering
    \fontsize{9}{11}\selectfont
    \setlength{\tabcolsep}{6pt}
    {\begin{tabular}{c|c|c}
    \toprule
    Label Smoothing Noise $\epsilon$ & ID& OOD\\
    \midrule
     0 &  77.3&  53.9\\
    0.01 & 77.5& 54.1\\
    0.03 & 77.5& 54.2\\
    \ccol {0.05} & \ccol \textbf{77.7}& \ccol \textbf{54.3}\\
    \bottomrule
    \end{tabular}}
\label{tab:eps_ablation}
\end{table}

\noindent \textbf{Ablation Study on Label Smoothing Coefficient.}
We conducted an ablation study on the label smoothing coefficient $\epsilon$, which is not included in the main text of the paper due to space limitations. The results of experiments on ImageNet using ViT-B/32 are presented in Table~\ref{tab:eps_ablation}. We observe that increasing the label smoothing parameter up to 0.05 leads to performance improvements in both In-Distribution (ID) and Out-of-Distribution (OOD) settings. However, we also notice that label smoothing does not always benefit all tasks. While there is a clear performance improvement in the full-shot setting of ImageNet classification, in cases with fewer samples like the few-shot setting, or in settings other than classification, even a weak label smoothing noise can deteriorate performance.
Our proposed loss, MPM-NCE, can consider multiple positive samples and also easily apply traditional regularization techniques like label smoothing, and thus get benefit from them.

\section{Training Time Comparison}

We compare and discuss the training latency of our method with the existing state-of-the-art method, Mask-Fill. The training latency for Mask-Fill is \textbf{8.44ms per image}, whereas, for our method, it is only \textbf{1.82ms per image}, tested on 64 batches with 3090 GPU. The training latency for Mask-Fill is computed using its official implementation\footnote[1]{\url{https://github.com/Coxy7/robust-finetuning}}. The reasons for the increased latency during training time and discussion comparing with our method are as follows:

Mask-Fill enhances robustness by using masked images as counterfactual samples, which helps improve the robustness of the fine-tuning model. It generates masked images and then distills the information for the masked parts from a pre-trained model. This process involves extra computation time for creating masks and generating new images by combining different images. Moreover, for distillation, two images need to be forwarded by the training model, and one of them is forwarded by a pre-trained model during each iteration. Consequently, this training method results in longer time consumption compared to conventional fine-tuning methods. In contrast, our method avoids such complex processing and learns fewer parameters, enabling faster training speeds. This experiment demonstrates that our method not only surpasses the existing state-of-the-art method in performance but is also superior in terms of training time.

\end{document}